\documentclass[12pt, a4paper]{article}

\usepackage[utf8]{inputenc}
\usepackage{geometry}
\usepackage{graphicx}
\usepackage{hyperref}
\hypersetup{
  colorlinks=true,
  linkcolor=blue,
  citecolor=blue,
  urlcolor=blue,
  breaklinks=true
}
\usepackage{xurl}
\usepackage{natbib}
\usepackage{authblk}
\usepackage{float}
\usepackage{subcaption}
\usepackage{tabularx}

\title{A Hierarchical Multi-Agent System for Autonomous Discovery in Geoscientific Data Archives}
\author[1]{Dmitrii Pantiukhin}
\author[1]{Ivan Kuznetsov}
\author[1]{Boris Shapkin}
\author[1]{Antonia Anna Jost}
\author[1]{Thomas Jung}
\author[1]{Nikolay Koldunov}
\affil[1]{Alfred Wegener Institute for Polar and Marine Research, Bremerhaven, Germany}
\date{}

\begin{document}

\maketitle

\begin{abstract}
The rapid accumulation of Earth science data has created a significant scalability challenge; while repositories like PANGAEA host vast collections of datasets, citation metrics indicate that a substantial portion remains underutilized, limiting data reusability. Here we present PANGAEA-GPT, a hierarchical multi-agent framework designed for autonomous data discovery and analysis. Unlike standard Large Language Model (LLM) wrappers, our architecture implements a centralized Supervisor-Worker topology with strict data-type-aware routing, sandboxed deterministic code execution, and self-correction via execution feedback, enabling agents to diagnose and resolve runtime errors. Through use-case scenarios spanning physical oceanography and ecology, we demonstrate the system's capacity to execute complex, multi-step workflows with minimal human intervention. This framework provides a methodology for querying and analyzing heterogeneous repository data through coordinated agent workflows.
\end{abstract}

\section{Introduction}
The increasing volume of digital data from sources such as satellite observations, autonomous underwater vehicles, and research vessels has increased the demand for scalable, automated analysis methods. Major international repositories, including PANGAEA, NASA's Earth Observing System Data and Information System (EOSDIS), and the Copernicus Climate Data Store (C3S), now support essential climate and oceanographic research, collectively hosting petabytes of observational and model-derived data \citep{kobler1995architecture, felden2023pangaea, li2023bigdata}. PANGAEA alone catalogs over 400,000 curated datasets spanning more than 800 distinct geoscientific parameters, from Conductivity-Temperature-Depth (CTD) hydrographic profiles and biogeochemical fluxes to ice-core proxy records and coupled model ensembles \citep{felden2023pangaea}. Yet the scale of curation exceeds the rate of scientific reuse: analysis of the Data Citation Index reveals that nearly 90\% of archived datasets have never been cited in the peer-reviewed literature \citep{robinson2016analyzing}.

This usability gap stems from interoperability challenges across the data lifecycle. Incomplete or inconsistent documentation obscures essential context (measurement provenance, instrument calibration history, quality-control flags), without which downstream reuse is impractical \citep{gil2016toward}. Format heterogeneity amplifies the problem: the same physical quantity may be encoded as a column in a flat CSV, a variable in a multi-dimensional NetCDF array, or a chunked Zarr store, each demanding a distinct software toolchain and non-trivial domain expertise to interpret correctly \citep{gil2016toward, schnase2016big}. While the FAIR (Findable, Accessible, Interoperable, Reusable) data principles provide an essential framework for addressing these barriers \citep{wilkinson2016fair}, implementing them at scale remains challenging even for dedicated community efforts \citep{marsden2025nansen}. The cumulative technical load required to discover, harmonize, and analyze disparate datasets remains a primary bottleneck to data reuse \citep{li2023bigdata}.

Large Language Models (LLMs) have evolved from text generators into autonomous agents that reason over problems, decompose tasks, and invoke external tools \citep{boiko2023emergent, schick2023toolformer}. This shift has given rise to Multi-Agent Systems (MAS) in which complex problems are partitioned across specialized agents \citep{hong2024metagpt, li2023camel, guo2024large}. Hierarchical agent architectures have demonstrated the ability to navigate knowledge graphs and design experiments in materials science \citep{li2024sciagents, chen2024atomagents}. Similarly, the AI Scientist-v2 \citep{yamada2025aiscientist} and Google's AI Co-Scientist \citep{gottweis2025coscientist} have been applied to automating research cycles and hypothesis generation, while AlphaEvolve \citep{novikov2025alphaevolve} uses LLM-driven evolutionary computation to discover and optimize algorithms, advancing results in mathematics and computing infrastructure. In laboratory settings, systems like Coscientist \citep{boiko2023autonomous} and ChemCrow \citep{bran2024chemcrow} have achieved autonomous planning in chemical synthesis. More recently, Robin \citep{aliseda2025robin} has automated hypothesis-driven biomedical discovery, Data Interpreter \citep{hong2025datainterpreter} targets general-purpose data science tasks, and DS-Agent \citep{guo2024dsagent} automates machine learning pipelines.

Current LLMs already rival human domain experts on graduate-level science benchmarks \citep{rein2024gpqa}, and their expanding context windows enable coherent multi-step tool use. Orchestration frameworks such as HuggingGPT show that an LLM controller can dynamically route subtasks to specialist models \citep{shen2023hugginggpt}. Whether these capabilities transfer to the messier reality of scientific data archives, with their inconsistent metadata and format heterogeneity, remains an open question.

Within geosciences, LLM adoption has progressed from domain-specific foundation models to tool-augmented autonomous agents. Models like GeoGalactica \citep{lin2023geogalactica} and OceanGPT \citep{bi2024oceangpt} established domain-specific linguistic capabilities. In the domain of climate services, ClimSight \citep{koldunov2024climsight, kuznetsov2025climsight} demonstrated that augmenting LLMs with localized climate model output, RAG-based report retrieval, and an agent-based architecture can deliver actionable, location-specific climate assessments at scale. Tool-augmented models like GeoGPT \citep{zhang2023geogpt} enabled geospatial data processing, followed by single-agent frameworks like GeoAgent \citep{chen2024llmagent} and LLM-Find \citep{ning2025autonomous}. However, as identified in our prior architectural survey \citep{pantiukhin2025pangaea}, none of these systems provide end-to-end integration with the retrieval, format-handling, and analytical infrastructure of a production-scale geoscientific data repository.

Despite this progress, existing approaches address individual components of the scientific workflow in isolation (discovery, analysis, or visualization) without integrating them into a unified, self-correcting pipeline capable of operating across the format diversity of a major repository. Moreover, no multi-agent system has been deployed for geoscientific data archives, where the inherent disciplinary breadth, spanning physical oceanography, biogeochemistry, paleoclimatology, and ecology, demands specialist agents with distinct analytical toolchains. What is missing is an architecture that combines semantic search, data-type-aware task routing, and deterministic code execution in a single orchestrated framework.

Here, we describe PANGAEA-GPT, a hierarchical multi-agent system first outlined in \citet{pantiukhin2025pangaea}, and provide its full architecture along with the first scenario-driven evaluation on real research workflows. We demonstrate that by combining an iterative, self-refining agentic search with a supervisor-orchestrated team of specialist data agents, the system can execute complex, multi-step Earth science analyses with reduced manual intervention.

\section{Results}

\subsection{System overview}
\begin{figure}[t!]
    \centering
    \includegraphics[width=\linewidth]{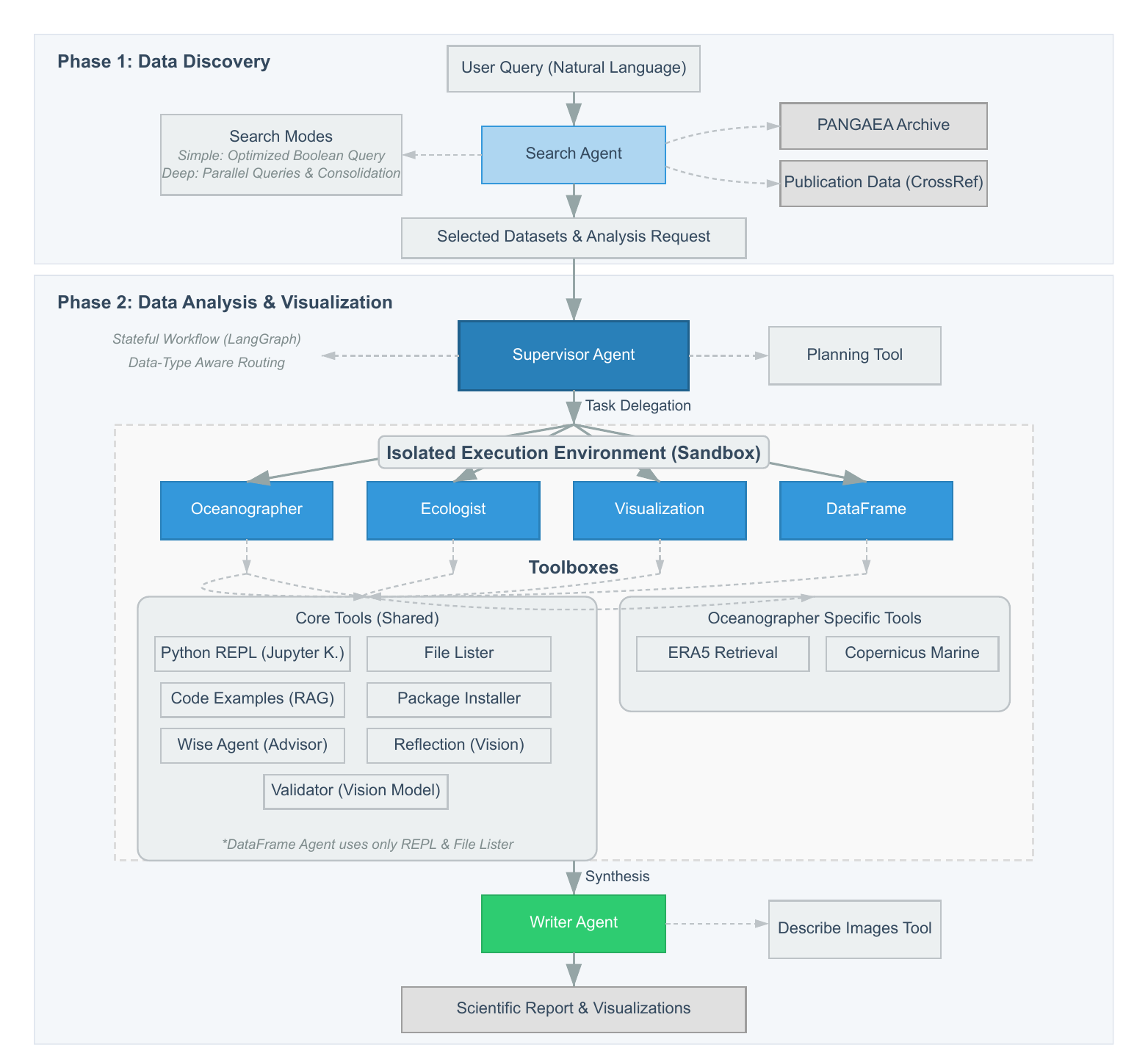}
    \caption{Conceptual framework of the PANGAEA-GPT Multi-Agent System (MAS). The system uses a two-phase, hierarchical architecture. First, a Search Agent discovers relevant datasets from a user's natural language request. A Supervisor Agent then delegates analysis and visualization tasks to a team of specialist agents working within a secure sandbox environment. Finally, a Writer Agent synthesizes the results into a cohesive report.}
    \label{fig:framework}
\end{figure}

The system operates through a hierarchical Supervisor-Worker topology that enforces a strict separation between reasoning and execution (Fig.~\ref{fig:framework}). A natural language query is received directly by the Supervisor Agent, which decomposes it into subtasks based on the schema, dimensionality, and coordinate reference system of each retrieved dataset. The Supervisor then routes each subtask to one of five specialist worker agents: (i)~an Oceanographer Agent for $N$-dimensional arrays (NetCDF, Zarr) and external climate reanalysis APIs; (ii)~an Ecologist Agent for biodiversity metrics and taxonomic aggregation; (iii)~a Visualization Agent with a reflexive quality-control loop and a retrieval-augmented generation (RAG) index of domain-specific exemplar plots; (iv)~a DataFrame Agent for statistical computation on tabular data; and (v)~a Writer Agent for narrative synthesis via a vision-language model interface (Methods~\ref{sec:routing}). This modality-based routing ensures that, for example, gridded arrays are never processed by tabular agents.

\subsection{Validation scenarios}
We evaluated PANGAEA-GPT on four scenarios that collectively test the full scientific workflow: data retrieval, cross-domain integration, statistical analysis, and visualization.

\subsubsection{Scenario 1: Microplastic distribution in the Weddell Sea}

Quantifying the coupling between ocean circulation and plastic accumulation typically requires bridging two distinct data regimes: Lagrangian point observations and Eulerian flow fields. To test the system's capacity for this cross-domain integration, we loaded a published dataset of sea-surface microplastic concentrations from the Weddell Sea ($N=34$ Manta trawl stations; DOI: \href{https://doi.org/10.1594/PANGAEA.941076}{10.1594/PANGAEA.941076}) and directed the agents to integrate it with ocean reanalysis fields.

\begin{figure}[t!]
    \centering
    \begin{subfigure}[b]{0.62\linewidth}
        \includegraphics[width=\linewidth]{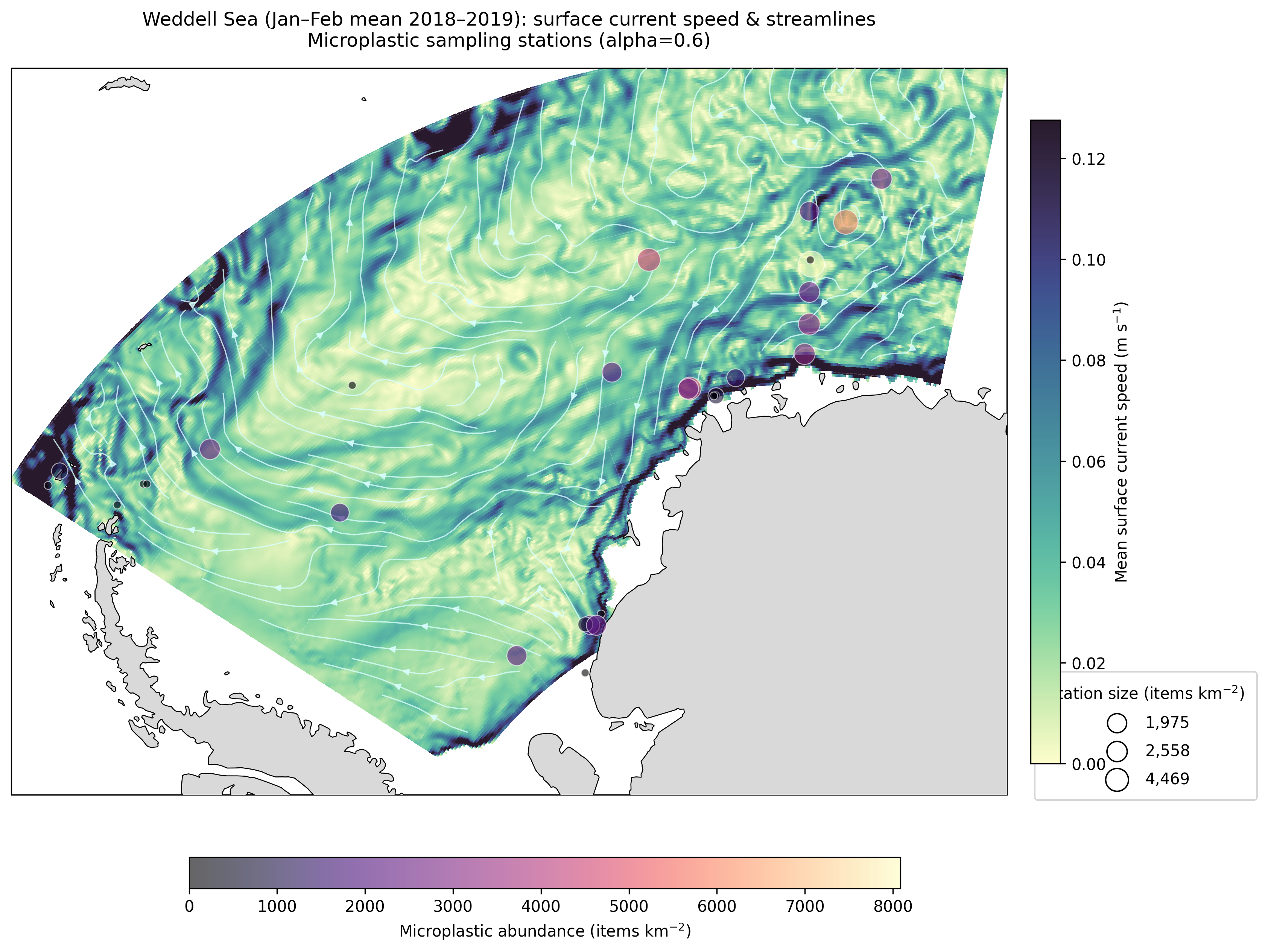}
        \caption{Mean surface current speed (January--February 2018--2019, GLORYS12V1) with streamlines and Manta trawl stations scaled by microplastic abundance. The visualization was iteratively refined via the reflexive quality control loop (3/10~$\rightarrow$~9/10).}
        \label{fig:gyre_map}
    \end{subfigure}\hfill
    \begin{subfigure}[b]{0.37\linewidth}
        \includegraphics[width=\linewidth]{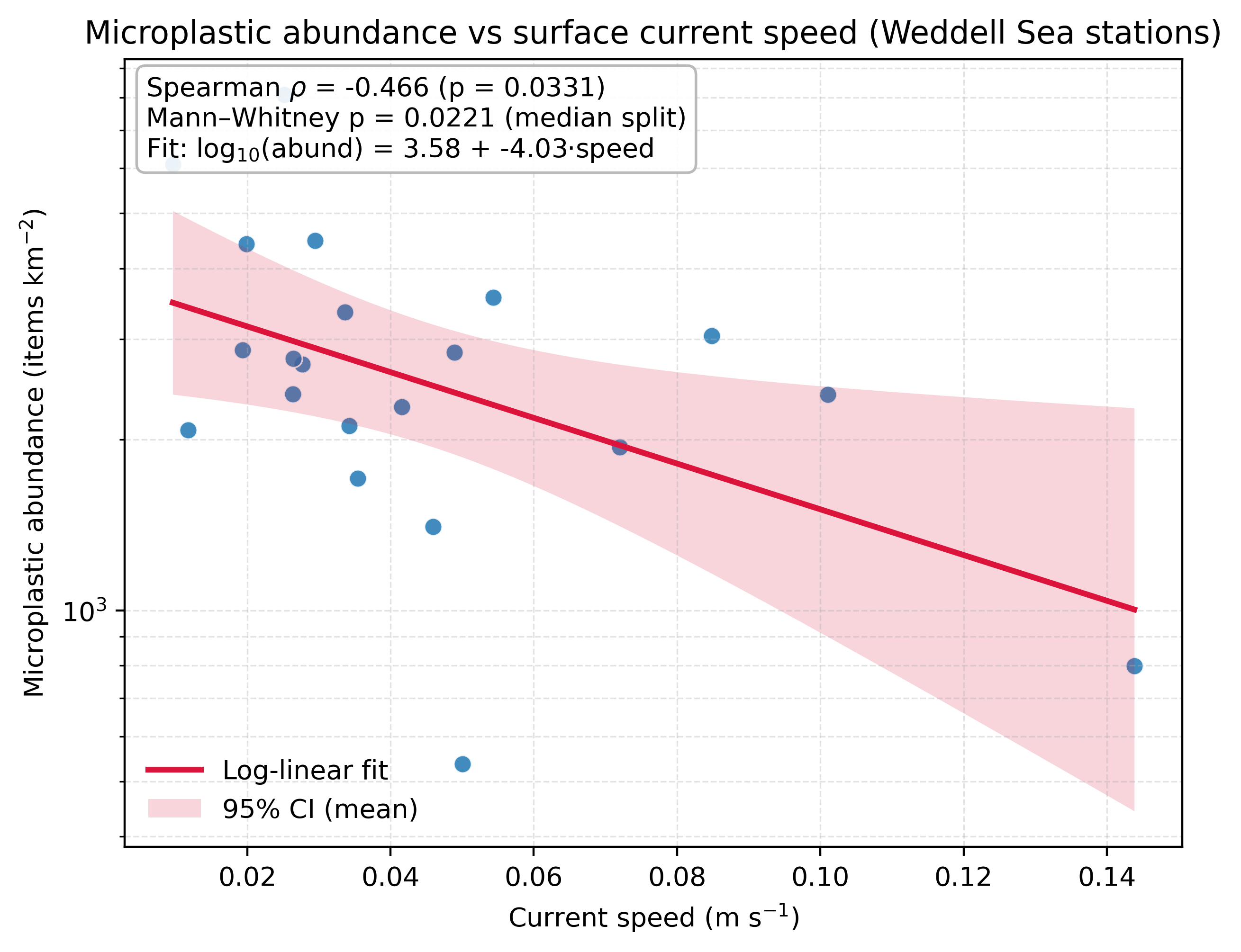}
        \caption{Log-linear regression of abundance against current speed ($N=21$). Annotations: Spearman $\rho=-0.47$ ($p=0.033$); Mann--Whitney $p=0.022$. Statistical annotations were automatically generated and placed on the plot by the Visualization Agent.}
        \label{fig:gyre_stats}
    \end{subfigure}
    \caption{\textbf{Exploratory analysis of microplastic distribution in the Weddell Sea} (Supplementary Note~1).}
    \label{fig:gyre}
\end{figure}

Guided by the user's definition of the study period, the Oceanographer Agent resolved the specific spatiotemporal constraints, retrieved daily surface velocity fields from the Copernicus Marine Service (GLORYS12V1), and computed a temporal mean circulation field over the months of the expedition. The Visualization Agent rendered a regional map (Fig.~\ref{fig:gyre_map}) overlaying current speed, flow streamlines, and stations scaled by microplastic abundance. The generated streamlines visually captured the cyclonic rotation of the Weddell Gyre. During execution, the agents autonomously resolved one API-level error and one visualization layout failure (Supplementary Table~S1). The system also handled domain-specific constraints of the underlying ocean model. When the initial request for surface-level data was rejected by the service, which defines its shallowest vertical layer at 0.494~m rather than the sea surface, the Oceanographer Agent autonomously parsed the error response, identified the nearest valid vertical coordinate, and resubmitted the retrieval without human intervention. In the visualization workflow, the reflexive critic identified that the legend obscured the central streamlines, prompting the agent to refactor the figure layout into dedicated colorbar and legend axes.

Following a user request to quantify the coupling, the DataFrame Agent performed a statistical evaluation on $N=21$ co-located stations (positive abundance, valid current speed), revealing a moderate negative association between current speed and microplastic concentration (Spearman $\rho = -0.47$, $p = 0.033$; Fig.~\ref{fig:gyre_stats}). A Mann--Whitney $U$ test confirmed significantly higher abundance in the low-speed regime ($U = 88.0$, $p = 0.022$; median 2,558 vs.\ 268 items\,km$^{-2}$). We emphasize that these statistics serve to demonstrate the system's capacity to autonomously execute a complete analytical pipeline, from data retrieval through statistical testing, rather than to draw robust ecological conclusions. With only 21 stations and no control for confounding variables, these results are exploratory and should not be treated as confirmatory.

\subsubsection{Scenario 2: Reanalysis validation against mooring observations}

Validating global climate models against in-situ observations typically requires extensive manual coding to bridge the gap between disparate tabular data and multi-dimensional gridded outputs. We tasked the system with validating temperature records from ten mooring datasets deployed at the LTER HAUSGARTEN observatory (Fram Strait, 2006--2012; PANGAEA DOIs 845608--845617) against the Copernicus Marine Service global reanalysis (GLORYS12V1).

\begin{figure}[t!]
    \centering
    \includegraphics[width=0.7\linewidth]{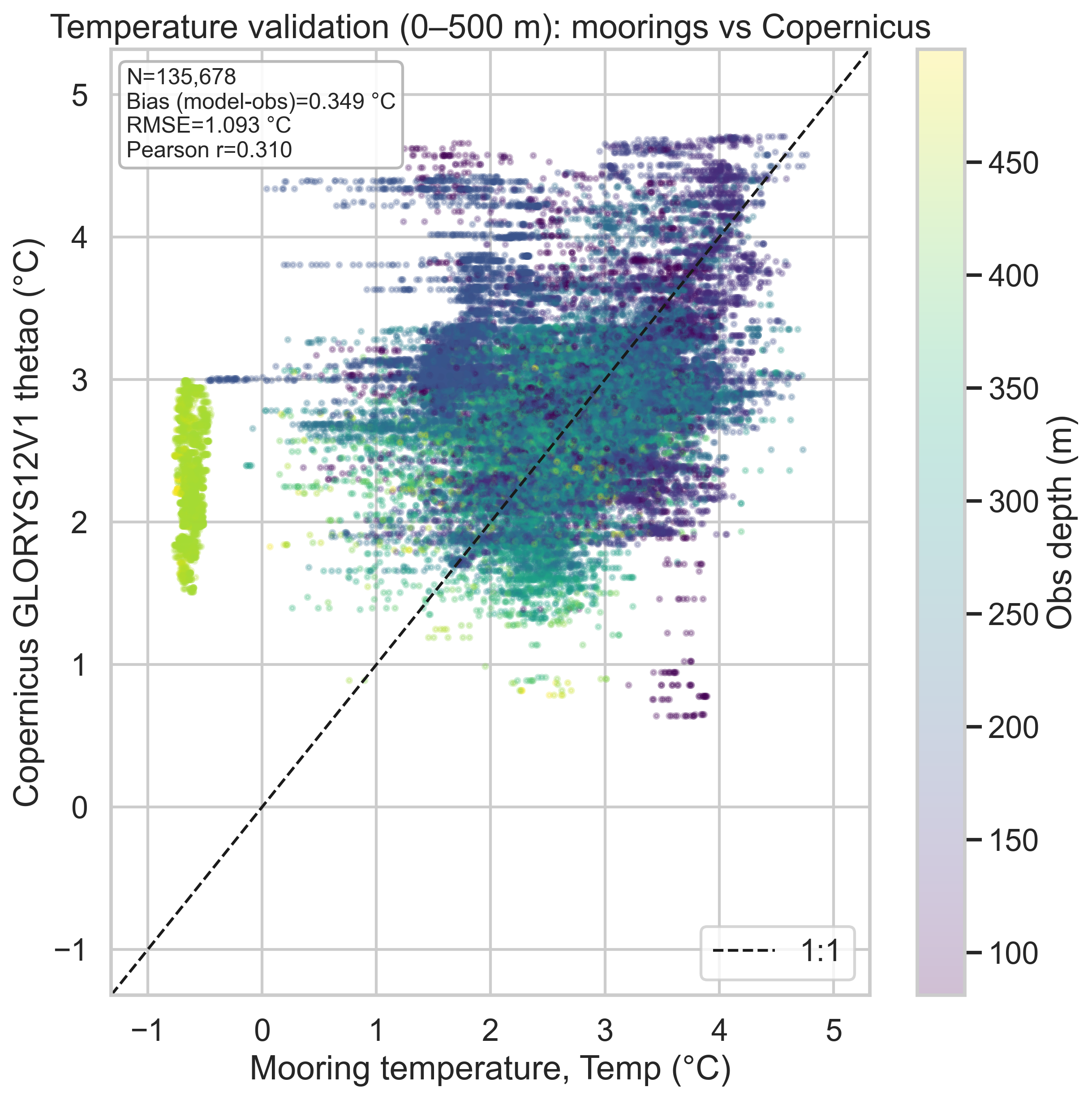}
    \caption{Validation scatter plot comparing in-situ temperature observations (0--500\,m) from ten HAUSGARTEN moorings with co-located values from the Copernicus GLORYS12V1 reanalysis. The Oceanographer Agent retrieved daily \texttt{thetao} fields, performed 4D nearest-grid-point matchup ($N=135{,}678$), and calculated statistics (Bias~$+0.35\,^\circ$C, RMSE~$1.09\,^\circ$C, $r=0.31$). Color scale indicates observation depth.}
    \label{fig:validation}
\end{figure}

The workflow proceeded through three stages. First, the Visualization Agent generated exploratory depth-coverage plots. Second, the DataFrame Agent programmatically harmonized the ten CSV files, standardizing column names and filtering for upper-ocean records ($N=135{,}678$). Finally, the Oceanographer Agent performed the 4D matchup. During this step, it encountered a type mismatch between the mooring timestamps and the reanalysis model's native date format. The agent diagnosed the incompatibility, refactored the temporal alignment, and re-executed the matchup, completing all 135,678 co-locations without human intervention (Supplementary Note~2). The resulting validation scatter plot (Fig.~\ref{fig:validation}) shows a warm bias of $+0.35\,^\circ$C, root-mean-square error (RMSE) of $1.09\,^\circ$C, and Pearson $r=0.31$. The positive bias is consistent with known offsets in the GLORYS12V1 reanalysis that are amplified in regions of intense Atlantic Water inflow such as the Fram Strait \citep{lellouche2021glorys}, and the moderate correlation reflects the use of nearest-grid-point matching without spatial interpolation across steep lateral temperature gradients between Atlantic and Arctic water masses.

\subsubsection{Scenario 3: Resource-efficient retrieval and trajectory analysis}

Validating gridded reanalysis products against moving platforms presents challenges regarding data transfer efficiency and geometric mismatch. We tasked the system with comparing ERA5 2-meter air temperature against shipboard measurements along the RV~Polarstern track during the MOSAiC expedition (PS122/1, September--December 2019; DOI: \href{https://doi.org/10.1594/PANGAEA.935221}{10.1594/PANGAEA.935221}).

\begin{figure}[t!]
    \centering
    \begin{subfigure}[b]{0.62\linewidth}
        \includegraphics[width=\linewidth]{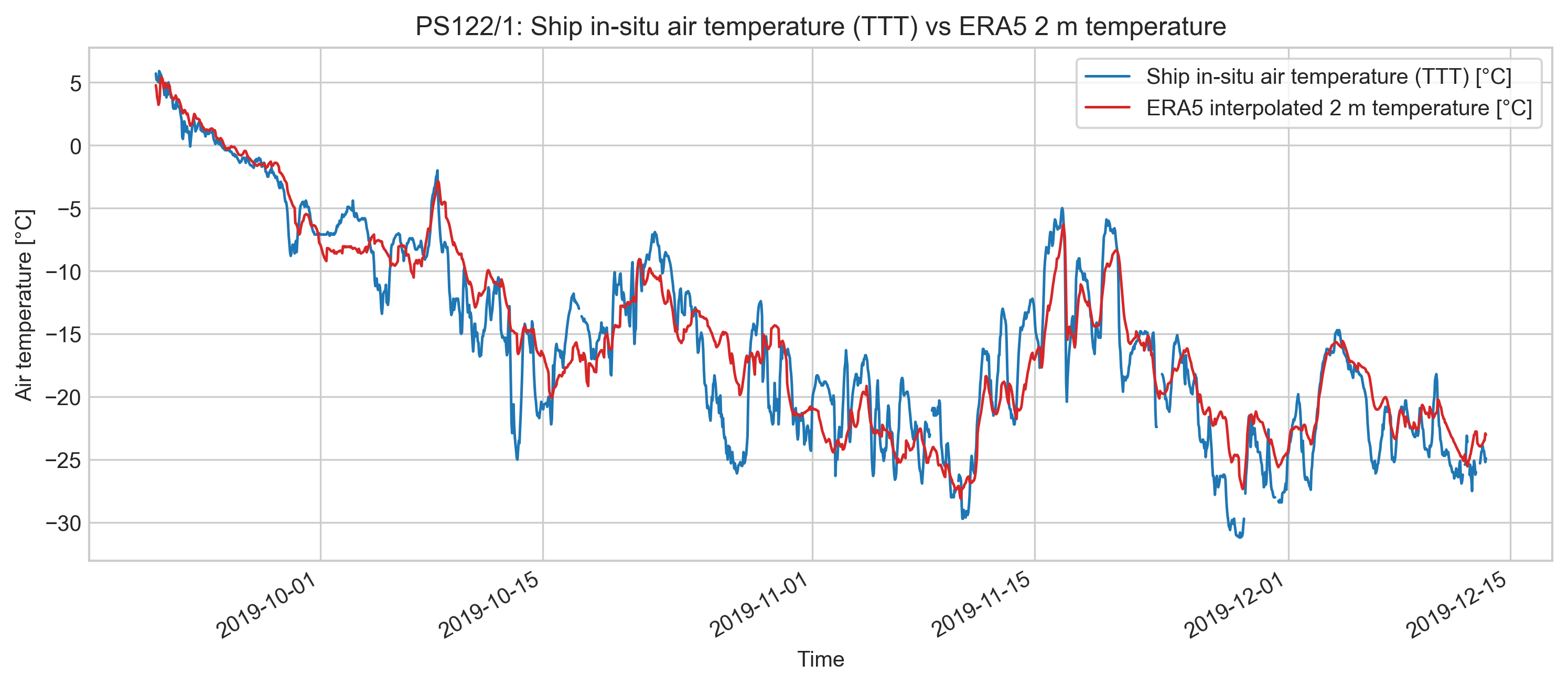}
        \caption{In-situ air temperature (blue) measured aboard RV~Polarstern compared with co-located ERA5 2-meter temperature (red) interpolated to the ship's hourly positions. The system calculated the bounding box to minimize cloud data retrieval (${\sim}$260\,MB).}
        \label{fig:lagrangian}
    \end{subfigure}\hfill
    \begin{subfigure}[b]{0.37\linewidth}
        \includegraphics[width=\linewidth]{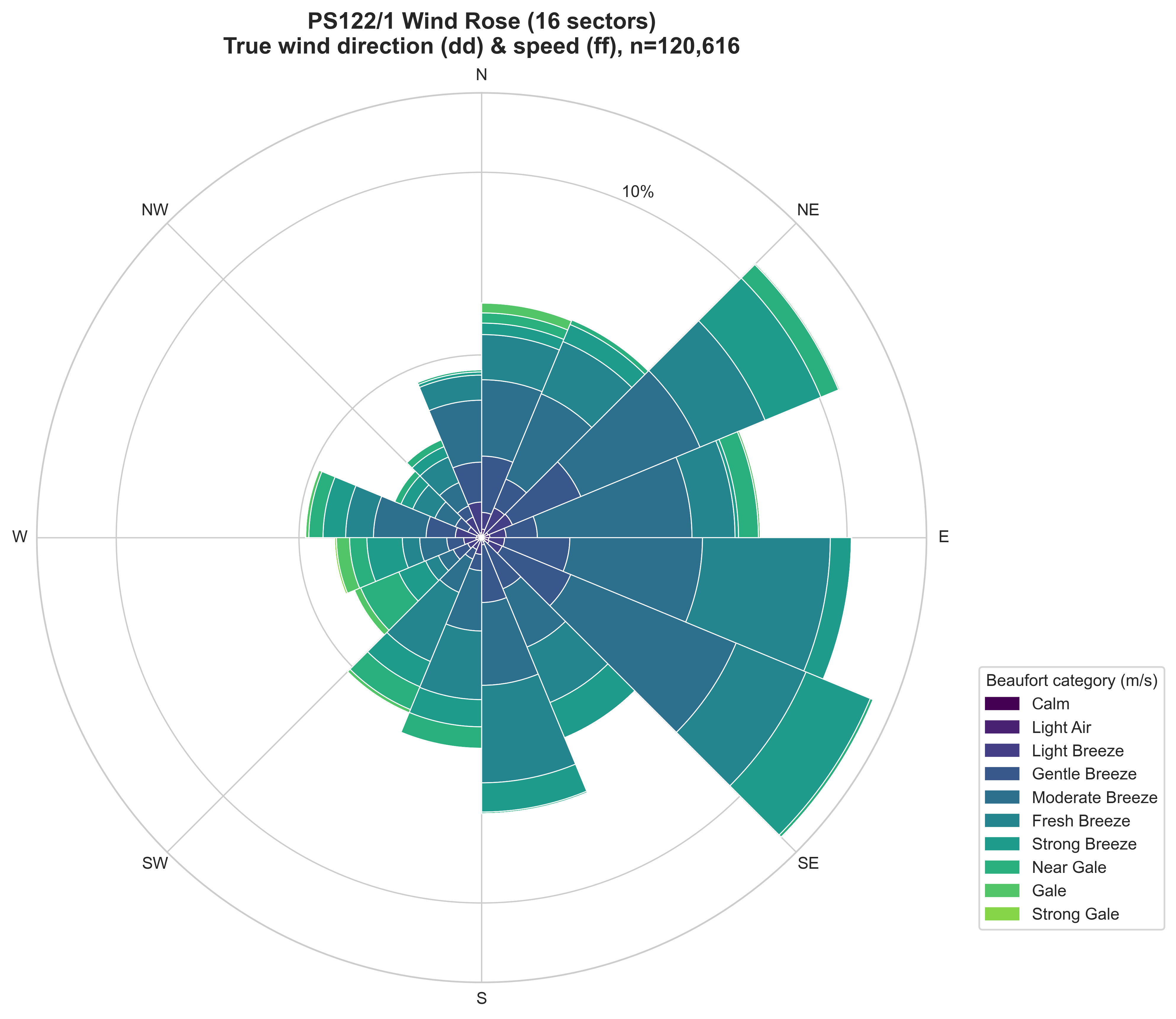}
        \caption{Wind rose diagram for the PS122/1 cruise leg (September--December 2019, $n=120{,}616$ observations). Wind directions are binned into 16 sectors; color scale indicates Beaufort wind force categories. The diagram was produced after one self-correction cycle.}
        \label{fig:windrose}
    \end{subfigure}
    \caption{\textbf{ERA5/MOSAiC Lagrangian validation and wind regime characterization.}}
    \label{fig:era5_mosaic}
\end{figure}

The DataFrame Agent parsed the navigation log to extract spatiotemporal bounds, which the Oceanographer Agent used to execute a targeted cloud-native retrieval from the Arraylake ERA5 data store \citep{earthmover2025arraylake}. The agent then interpolated the gridded field to the ship's 2,013 hourly positions, producing a co-located time-series (Fig.~\ref{fig:lagrangian}). A subsequent prompt directed the system to generate a wind rose diagram (Fig.~\ref{fig:windrose}). The agents recovered from two runtime errors, including an out-of-bounds interpolation edge case (Supplementary Note~3). During interpolation of the ship track against the gridded ERA5 field, the Oceanographer Agent detected that 83 of the 2,013 co-located points returned invalid values where the vessel trajectory grazed the bounding-box edge. Rather than reporting a partial result, the agent autonomously expanded the retrieval domain by approximately 0.3$^\circ$ and re-executed the extraction, achieving 100\% spatial coverage. The wind rose generation required the agent to autonomously diagnose a dimensional mismatch between the number of speed-bin edges and the corresponding Beaufort scale labels (an off-by-one error inherent in binned classification) and correct the bin definitions before producing the final diagram.

\subsubsection{Scenario 4: Biodiversity--water mass coupling}

Interdisciplinary research often requires merging sparse biological observations with continuous physical fields. We tasked the system to compare jellyfish biodiversity between the Norwegian and Irminger Seas using a cruise dataset from the North Atlantic (May 2013; DOI: \href{https://doi.org/10.1594/PANGAEA.829702}{10.1594/PANGAEA.829702}) and to link species abundance to specific water masses.

First, the Ecologist Agent stratified sampling stations into ``Eastern'' (Norwegian Sea) and ``Western'' (Irminger Sea) transects based on a user-defined $20^\circ$W longitude threshold. It then computed the Shannon-Wiener diversity index ($H'$) for each station. A Welch's t-test ($t=-3.39$, $p=0.0014$) indicated significantly higher biodiversity in the Western transect ($\bar{H'}_{west}=0.81$) compared to the Eastern transect ($\bar{H'}_{east}=0.35$). During the diversity computation, the Ecologist Agent encountered a division-by-zero error caused by zero-abundance taxa. It autonomously applied a masking operation to exclude these entries before the logarithmic transformation. The Visualization Agent produced a comparative violin plot, utilizing the reflexive quality control loop to detect and fix label overlaps across five iterations.

\begin{figure}[t!]
    \centering
    \includegraphics[width=0.7\linewidth]{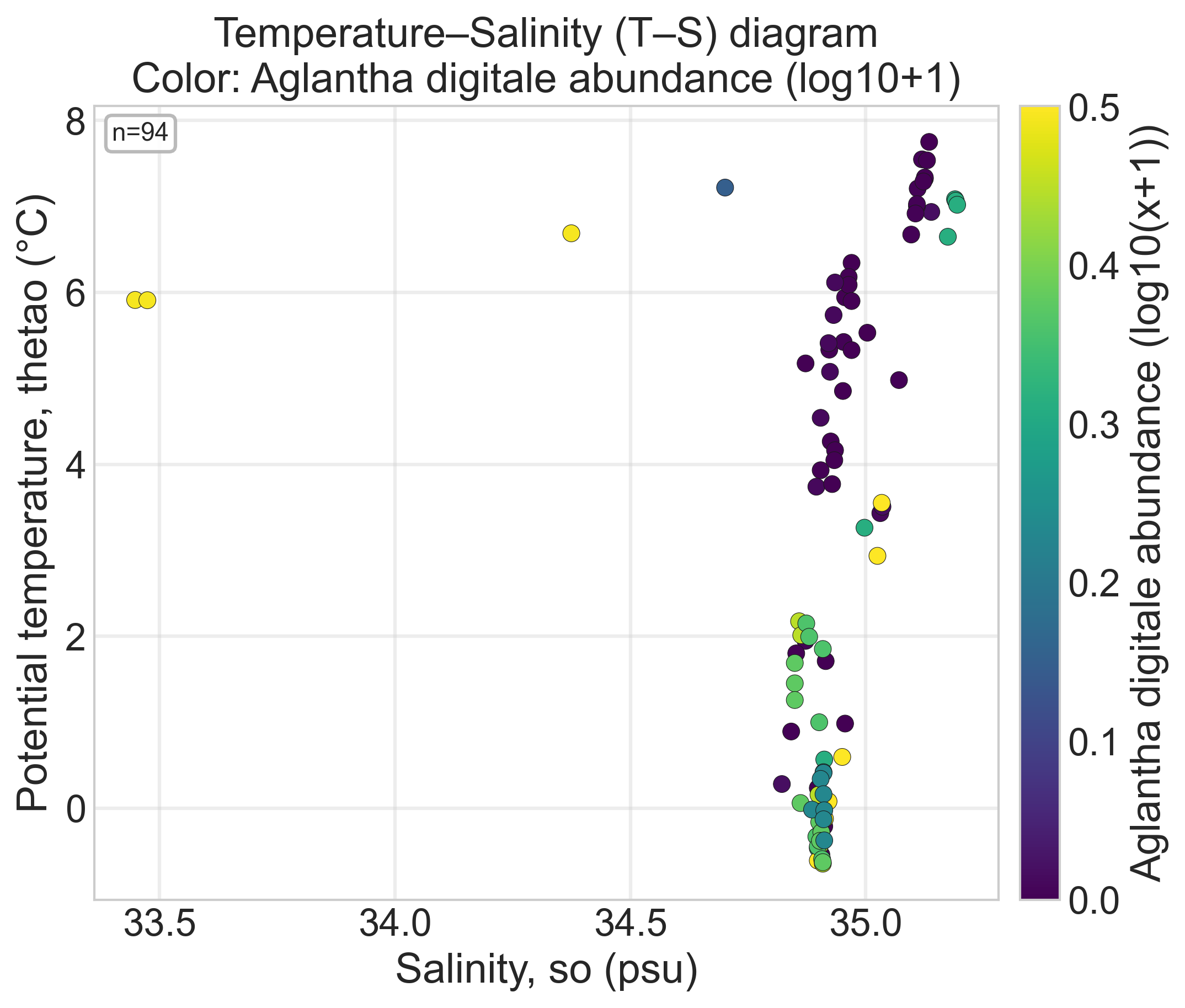}
    \caption{\textbf{Bio-physical coupling generated by the Oceanographer Agent.} The system augmented the PANGAEA biological dataset by retrieving and co-locating 4D hydrographic data ($\theta$/$S$) from the Copernicus Marine Service GLORYS12V1 reanalysis. The T-S diagram highlights the co-occurrence of \textit{Aglantha digitale} with distinct water masses ($\log_{10}$-abundance).}
    \label{fig:ts_diagram}
\end{figure}

Subsequently, to determine the environmental niche of the dominant hydrozoan \textit{Aglantha digitale}, the Oceanographer Agent queried the Copernicus Marine Service to retrieve co-located potential temperature ($\theta$) and salinity ($S$) profiles. It performed a 4D nearest-neighbor matchup between the global reanalysis model and the local biological samples. The resulting Temperature-Salinity (T-S) diagram (Fig.~\ref{fig:ts_diagram}) shows that the highest abundances of \textit{A.~digitale} cluster within the cold, fresh end-member ($\theta < 2\,^\circ$C, $S \approx 34.8$--$35.0$\,psu), corresponding to Subpolar Mode Water carried by the Irminger Current. This is consistent with the species' known affinity for subarctic water masses in the North Atlantic \citep{pantiukhin2024panarctic}. The warmer, more saline Norwegian Atlantic Current stations ($\theta > 5\,^\circ$C, $S > 35.1$\,psu) hosted markedly lower abundances (Supplementary Note~4).

\subsection{Search benchmark}
To evaluate the agentic retrieval architecture, we conducted a benchmark using a curated dataset of 100 natural language queries spanning six geoscientific domains: physical oceanography, marine geology, biogeochemistry, paleoclimatology, marine ecology, and atmospheric science (Supplementary Note~5). Queries were manually assembled to represent realistic research needs and stratified into five complexity categories: single-parameter lookups, multi-constraint filtering, spatiotemporal subsetting, cross-domain integration, and open-ended discovery. Each query's retrieval results were evaluated using a structured rubric scored on a 1--10 scale across five metrics: Precision, Recall, Parameter Coverage, Metadata Quality, and Result Relevance. Evaluation was performed programmatically using Gemini 3 Pro Preview \citep{gemini2025} as an automated judge, following a structured rubric with detailed scoring criteria for each metric. All search architectures were driven by GPT-5.2 \citep{openai2025gpt52}. We compared three configurations: (1) Baseline (Elasticsearch BM25, raw keyword matching), (2) Simple LLM (single-pass translation of the natural language query into a structured Boolean request), and (3) Agentic Search (our ReAct-based iterative agent, which decomposes user intent into multiple Boolean query permutations and refines them across successive retrieval cycles).

\begin{figure}[t!]
    \centering
    \includegraphics[width=0.8\linewidth]{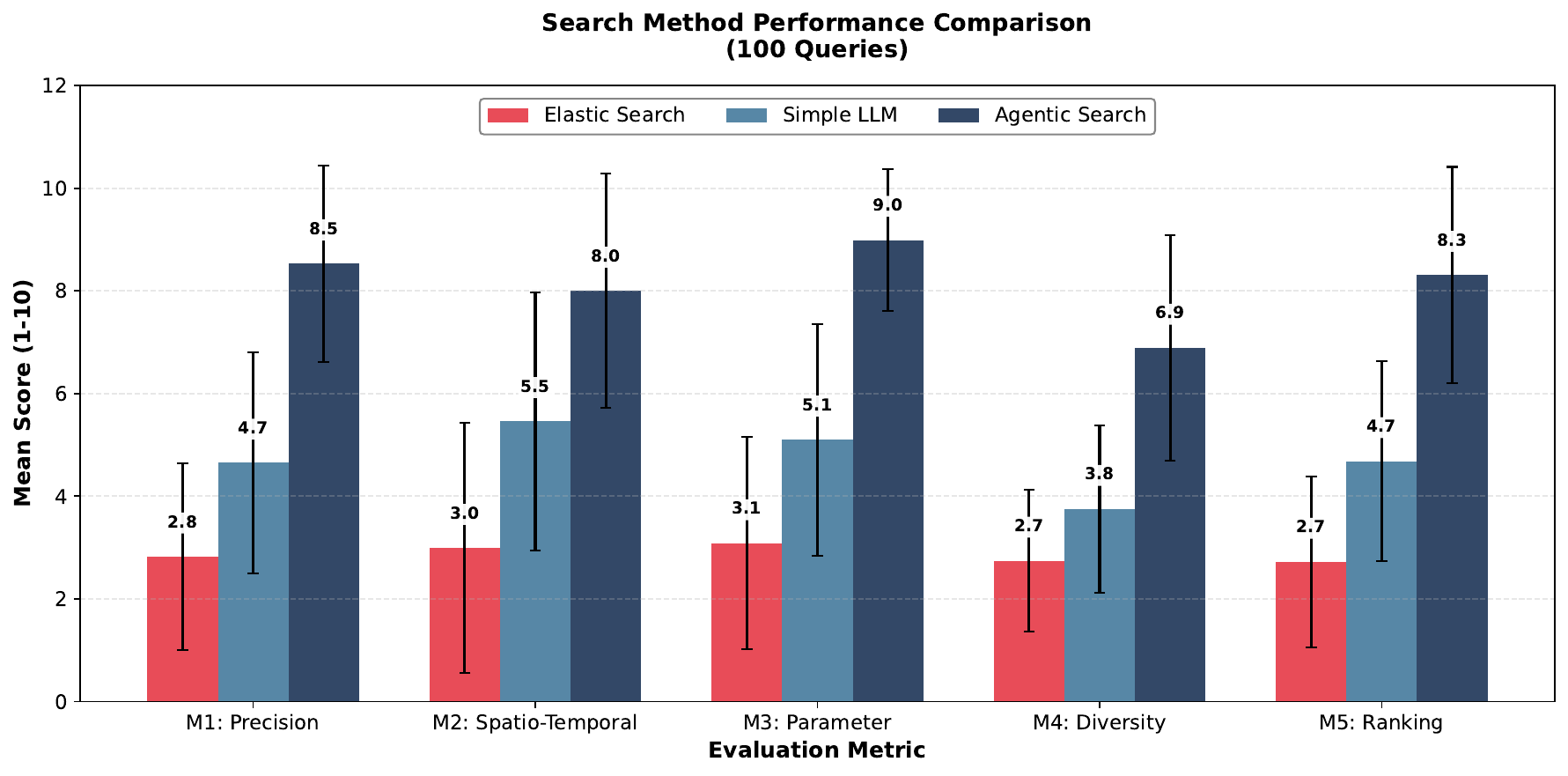}
    \caption{Grouped bar chart comparing Baseline (Elastic Search), Simple LLM, and Agentic Search architectures across five semantic metrics using 100 geoscientific queries. Error bars indicate standard deviation. The Agentic Search method outperforms traditional approaches in all categories.}
    \label{fig:benchmark}
\end{figure}

The Agentic Search strategy increased the overall mean score to 8.14/10 compared to the Simple LLM's 5.46/10 and the Baseline's 2.87/10 (Fig.~\ref{fig:benchmark}). The most significant gains occurred in Precision (M1: 8.53) and Parameter Coverage (M3: 8.99). Beyond aggregate statistics, qualitative analysis of individual queries reveals distinct failure modes that the agentic architecture resolves (Supplementary Note~6).

\textbf{Temporal reasoning.} When queried for ``salinity profiles from the Weddell Sea during winter 2013,'' both the Baseline and Simple LLM failed to resolve the hemispheric seasonal constraint: the former returned IceBird altimetry campaigns from the wrong year, while the latter selected summer cruises. Only the Agentic Search correctly mapped ``Austral Winter'' to June--August and ranked the ANT-XXIX/6 (AWECS) cruise as the top result (Agentic: 9.8/10 vs.\ Baseline: 1.8/10; full analysis in Supplementary Note~6).

\textbf{Source disambiguation.} A query for ``satellite altimetry data'' exposed a systematic retrieval error: both the Baseline and Simple LLM returned in-situ mooring datasets whose metadata mentioned ``used to validate satellite altimetry,'' confusing ground-truth with the requested product. The Agentic Search correctly prioritized gridded sea-surface height anomaly products from CryoSat-2, Envisat, and TOPEX/Poseidon (Agentic: 9.6/10 vs.\ Baseline: 1.0/10).

These examples indicate that iterative query refinement resolves implicit scientific constraints (hemispheric seasonality, measurement geometry, product-level intent) that keyword and single-pass approaches systematically miss.

\section{Discussion}

Four architectural primitives together enable the composition of multi-step geoscientific workflows across heterogeneous data formats: data-type-aware routing, sandboxed execution, self-correction from execution tracebacks, and reflexive visual quality control. Iterative agentic search treats user queries not as strings to be matched but as under-specified scientific intents to be progressively refined. These primitives generalize across the validation scenarios without scenario-specific engineering, and by resolving implicit constraints that keyword and single-pass LLM approaches miss, the system reduces the manual effort required to transition from data retrieval to analysis \citep{singh2025agenticrag, zhang2023datacopilot}.

Agentic search works because it replicates what a domain expert does implicitly during manual retrieval. When a researcher searches for ``sediment cores from the Fram Strait,'' they do not submit these words verbatim; they translate the geographic name into a bounding-box specification, identify relevant campaign identifiers, and cross-reference parameter inventories. In practice, however, no user formulates this as a structured Boolean query against an Elasticsearch index; they type natural language and expect relevant results. The LLM bridges this gap by translating free-text intent into structured Boolean queries, and the agentic layer goes further by generating multiple query permutations and iteratively refining them against the returned metadata. The four validation scenarios stress-test different capabilities of this approach: cross-domain data fusion (Scenario~1), format harmonization across heterogeneous tabular sources (Scenario~2), resource-efficient cloud-native retrieval along non-Eulerian trajectories (Scenario~3), and biological--physical coupling requiring external reanalysis integration (Scenario~4). We note that the system functions as an assisted analysis tool whose output quality depends on the specificity of the user's intent; vague or under-constrained queries may yield retrieval of marginally relevant datasets.

The layered self-correction strategy addresses error propagation at two levels. At the programmatic level, tracebacks from the sandboxed Python kernel are fed back to the generating agent, which parses the error, refactors the code, and re-executes. In Scenario~1, for example, the Oceanographer Agent diagnosed and corrected a Copernicus API rejection caused by an invalid depth coordinate. At the perceptual level, the visual quality-control loop subjects every generated figure to a structured critique. In Scenario~4, this loop drove the violin plot through five successive refinements (scores 3$\rightarrow$4$\rightarrow$4$\rightarrow$5$\rightarrow$9/10), each targeting a distinct rendering defect. Across all scenarios, the system resolved all five runtime errors without human intervention, and the visual QC loop executed 11 critique cycles across the generated figures (Supplementary Table~S6). This combination of programmatic and perceptual feedback draws on methods from verbal reinforcement learning \citep{du2024debate, shinn2023reflexion} and multi-agent collaborative verification \citep{darwish2025hallucination, shi2025mcf}. A third mechanism, cross-model escalation to a secondary foundation model (Section~\ref{sec:wise_agent}), provides an additional fallback, though it was not triggered in the four validation scenarios.

However, the centralization of task decomposition in the Supervisor Agent constitutes a single point of failure. If the Supervisor incorrectly decomposes a query (for example, routing a gridded reanalysis comparison to the DataFrame Agent instead of the Oceanographer Agent), downstream agents have no mechanism to detect or correct this misassignment. Similarly, the system cannot verify domain-specific correctness beyond code executability. For example, it cannot detect whether a co-located dataset contains a systematic spatial offset from an incorrect coordinate reference system, or whether an unnecessary unit conversion was applied. Addressing these failure modes will require domain-specific validation hooks, such as automated CRS consistency checks and unit-conversion verification, that operate independently of the Supervisor's planning logic.

The Supervisor-Worker topology also facilitates reproducibility and extensibility. Each analytical step is captured as executable Python code within a deterministic sandbox, providing a complete provenance trail that complements the stochastic reasoning layer \citep{herschel2017provenance}. Extending the system to new domains requires three steps: defining a domain-specific agent prompt, registering routing rules that map feature flags to the new agent, and adding the required feature flags to the metadata introspection module (Section~\ref{sec:metadata_enrichment}). Natural candidates include sedimentology (grain-size distributions, stratigraphic correlation), paleoclimatology (proxy calibration, age--depth modelling), and geochemistry (elemental ratios, isotope systematics). This modular design follows principles established in prior multi-agent frameworks \citep{shen2023hugginggpt, hong2024metagpt}.

The current system operates reactively, responding to individual user queries. However, the architecture is compatible with an asynchronous mode in which agents traverse the archive over extended periods, systematically loading datasets, cross-referencing observations against reanalysis fields, and flagging statistical anomalies, metadata inconsistencies, or previously unexamined inter-dataset correlations. For example, an agent could compare all CTD profiles in a given region against the corresponding reanalysis grid cells, identifying systematic observation--model discrepancies that may indicate instrument calibration drift or ocean features not captured by the model. A complementary application is inverse metadata generation: agents could ingest poorly documented legacy files, infer parameter names and spatiotemporal coverage from data distributions and file headers, and produce machine-actionable metadata records, increasing the fraction of discoverable datasets \citep{jacobsen2020fair}. Such long-running exploration would generate candidate findings for subsequent human review, turning the archive from a passive store into one that actively surfaces candidate patterns for expert review. We note that this direction is speculative and would require advances in hallucination detection \citep{farquhar2024semantic}, automated significance filtering to suppress spurious correlations, and human-in-the-loop validation protocols before operational deployment.

Modularity is also a practical design requirement for systems of this kind. Because each worker agent communicates with the Supervisor only through a standardized message interface, individual agents can be detached from the pipeline and reused independently. The Oceanographer Agent, for example, already integrates ERA5 and Copernicus data without relying on the PANGAEA retrieval layer, and could serve as a standalone reanalysis tool in other workflows. Designing agents to be self-contained from the outset lowers the cost of adapting the system to new repositories or embedding individual components into third-party platforms.

\section{Limitations and Future Directions}

Our results must be interpreted within the constraints of the underlying data infrastructure. The system's retrieval performance is tied to the quality of existing metadata; in the PANGAEA archive, only approximately 15\% of datasets contain structured abstracts, and sparse parameter descriptions limit the efficacy of semantic search for the remaining majority. This creates a ``long-tail'' problem where poorly documented datasets remain effectively invisible to the retrieval layer. Additionally, while the architectural components are not specific to PANGAEA, our evaluation was restricted to this repository. Transferability to other major repositories, such as NASA EOSDIS or GBIF, remains untested and will require handling distinct API conventions, authentication mechanisms, and metadata schemas.

Because the system relies on stochastic inference, outputs are inherently non-deterministic, which conflicts with the reproducibility requirements of scientific computing. Repeated execution of the same query may yield functionally equivalent but syntactically different code, different dataset retrieval orderings, and marginally different numerical results due to alternative interpolation strategies or coordinate-matching heuristics. While the deterministic sandbox ensures that any given code block produces identical output, the upstream reasoning layer introduces variability in which code is generated.

The system also remains vulnerable to hallucinations at two levels \citep{huang2023hallucination}. Visual hallucinations (rendering artefacts, incorrect axis labels, missing legends) are mitigated by the reflexive quality-control loop described above. Interpretive hallucinations are more difficult to detect: the system might correctly compute a correlation coefficient but attribute it to a causal mechanism unsupported by the data, or generate a narrative that over-interprets a statistically marginal result. The system also shows reduced performance on multi-step reasoning that requires domain-specific knowledge; for example, inferring that a depth coordinate mismatch implies an incompatible vertical reference frame rather than a unit conversion error \citep{wang2024scibench}. As domain-specific foundation models mature, integrating them into the reasoning layer may reduce these failure modes.

Each multi-step analytical session involves dozens of LLM inference calls spanning task planning, code generation, error diagnosis, and visual critique, resulting in non-trivial computational costs when using commercial models such as GPT-5.2. Emerging open-weight models and model distillation techniques may reduce these costs, particularly for routine operations such as code refactoring and metadata parsing, enabling deployment on institutional computing clusters.

From a security perspective, although the sandbox provides filesystem and process isolation, the system does not currently perform static analysis or formal verification of agent-generated code. Future hardening should include AST-level code inspection, per-session resource quotas (CPU time, memory, disk), and an explicit allowlist of importable Python packages \citep{chen2025reproducibility}. In multi-tenant deployments, session isolation must also prevent cross-user data leakage through shared kernel state.

Finally, our evaluation methodology introduces a circularity: an LLM (Gemini 3 Pro Preview) serves as the automated judge for a system driven by another LLM (GPT-5.2). The judge evaluates factual alignment between returned datasets and query constraints but does not assess the reasoning process that led to dataset selection. While this approach enables reproducible scoring at scale, future validation must incorporate human expert annotation on a representative subset of queries to calibrate automated scores, quantify inter-rater agreement, and identify systematic blind spots in the automated evaluation. The full scoring rubric and per-query results are provided in Supplementary Note~5.

\section{Methods}

\subsection{Search and retrieval}
\label{sec:search_methods}

To bridge the semantic gap between natural language inquiry and the PANGAEA archive schema, we designed and evaluated a multi-tiered retrieval architecture comprising three configurations of increasing autonomy (Supplementary Note~5). Each tier was benchmarked against a curated dataset of 100 natural language queries spanning six geoscientific domains, scored by an automated judge on five semantic metrics (Supplementary Note~5).

\subsubsection{Baseline: Elasticsearch keyword retrieval}
The lowest tier relies on the PANGAEA Elasticsearch index using standard BM25 scoring with tf-idf weighting. Queries are matched verbatim against metadata fields (title, abstract, parameter names, campaign identifiers). This configuration provides sub-second latency but lacks semantic understanding: synonymous terms (e.g., ``CTD'' vs.\ ``conductivity--temperature--depth''), geographic colloquialisms (``Fram Strait'' vs.\ a bounding-box specification), and temporal abstractions (``Last Glacial Maximum'') are not resolved. The Baseline achieved a mean score of 2.87/10 in our benchmark, reflecting high recall for trivially specified queries but poor precision and parameter coverage for complex scientific requests.

\subsubsection{Intermediate: simple LLM query translation}
The intermediate tier interposes a single LLM inference call between the user and the search engine. GPT-5.2, operating in a zero-shot setting, translates the natural language query into a structured Lucene Boolean string, resolving geographic names to coordinate ranges and expanding parameter synonyms. This single-pass approach improved the benchmark mean to 5.46/10 but remains brittle: the model has no mechanism to inspect the returned results, detect low-quality matches, or iteratively refine the query. Ambiguous queries that require multiple rounds of disambiguation therefore saturate at moderate precision.

\subsubsection{Advanced: agentic search}
The Search Agent employs a ReAct (Reasoning + Acting) loop \citep{yao2023react}, operating through four stages:
\begin{enumerate}
    \item \textbf{Inferential Query Translation:} The agent decomposes user intent into metadata constraints, deriving implicit boundary conditions where needed; for example, translating ``Fram Strait'' or ``Last Glacial Maximum'' into bounding-box coordinates and numeric temporal intervals.
    \item \textbf{Execution and Introspection:} The agent executes a structured query and evaluates the returned metadata.
    \item \textbf{Parallelized Query Expansion:} Rather than relying on a single search string, the agent programmatically generates a matrix of Boolean query permutations targeting distinct metadata fields, permuting instrument types, parameter synonyms, and platform identifiers. These queries are executed concurrently against the archive.
    \item \textbf{Semantic Consolidation:} The system aggregates results, performs deduplication based on unique identifiers, and ranks each result by scoring its metadata against predefined relevance criteria derived from the original user prompt.
\end{enumerate}

\subsubsection{Metadata retrieval and feature introspection}
\label{sec:metadata_enrichment}
Before any analytical code is generated, the system pre-fetches the full metadata record for each selected dataset, including parameter inventories, matrix dimensions, coordinate reference systems, and temporal extents. From these records, it derives binary feature flags that characterize each dataset along five dimensions: whether it contains geospatial coordinates, whether it is stored as a gridded array, whether it includes a depth axis, whether its file size exceeds a configurable threshold, and whether it carries a temporal index. The system injects these flags into the Supervisor's planning context, where they directly determine the routing decision (Section~\ref{sec:routing}). For instance, a dataset flagged as gridded with a depth axis is routed to the Oceanographer Agent, whereas a flat table without spatial coordinates is assigned to the DataFrame Agent. This pre-execution introspection eliminates the ambiguity that would otherwise arise if the Supervisor inferred data type from the user's natural language description alone.

\subsection{Multi-agent orchestration}
The analytical core is a stateful MAS governed by a directed cyclic graph, implemented using the LangGraph framework \citep{langgraph2024} built on top of LangChain \citep{langchain2024}, with a Supervisor Agent as the orchestrator. The architecture isolates the reasoning engine from the computational kernel, enabling task routing based on data-type introspection and ensuring reproducibility through deterministic code execution.

\subsubsection{Supervisor agent and task planning}
\label{sec:supervisor}
The Supervisor maintains the global conversational state and manages inter-agent handovers. Before delegating tasks, it inspects the internal structure of retrieved resources (e.g., NetCDF dimension hierarchies, Zarr chunk layouts) to select the appropriate specialist. It decomposes user queries into sub-tasks via a planning module, dispatches them sequentially, and updates the plan as results become available. Safeguards include an iteration cap and strict data-type routing constraints to prevent, for instance, routing multi-dimensional arrays to agents specialized for tabular data.

\subsubsection{Data-type routing and agent specialization}
\label{sec:routing}
The introspection performed during the metadata enrichment phase (Section~\ref{sec:metadata_enrichment}) determines the execution branch, routing tasks to specific worker nodes based on their architectural specialization:
\begin{itemize}
    \item \textbf{The Oceanographer Agent:} Specialized for $N$-dimensional array handling (NetCDF/Zarr). It possesses tool-augmented access to external climate reanalysis APIs (Copernicus Marine Service, ERA5 via Arraylake~\citep{earthmover2025arraylake}), enabling contextualization of local observations within global environmental fields. The agent automatically extracts spatiotemporal bounds from loaded datasets, retrieves the matching reanalysis subset via lazy-loaded cloud-native Zarr stores, and performs geometric co-location (nearest-grid-point for Eulerian comparisons, 4D interpolation for Lagrangian trajectories). Full details are provided in Supplementary Note~7.
    \item \textbf{The Ecologist Agent:} Optimized for biodiversity metrics and tabular data manipulation, equipped with specialized libraries for biological indices calculation.
    \item \textbf{The Visualization Agent:} A general-purpose plotting specialist augmented with a Retrieval-Augmented Generation (RAG) index of domain-specific exemplar plots, enabling it to match figure types to data dimensionality and enforce scientific conventions (e.g., reversed depth axes, standard color palettes).
    \item \textbf{The DataFrame Agent:} Strictly confined to data manipulation, statistical analysis, and filtering of tabular data. This agent is architecturally prohibited from generating visualizations, enforcing a clean separation between computation and presentation.
    \item \textbf{The Writer Agent:} A synthesis specialist that collects outputs from upstream agents and produces a coherent scientific narrative, interpreting generated figures via a vision-language model interface.
\end{itemize}

\subsubsection{Adaptive context management}
\label{sec:memory}
Building on the multi-tier memory concept described in \citet{pantiukhin2025pangaea}, we implement a progressive summarization strategy to manage context window limits. The system partitions conversational history into two segments: recent messages retained at full resolution, and older exchanges compressed into a structured summary by a fast inference model. This summary explicitly preserves critical operational state (variable names, loaded dataset identifiers, file paths, and intermediate results) while discarding conversational filler. The result is that agents retain sufficient context to resume multi-step workflows even when the raw token history exceeds the model's context window.

\subsubsection{Sandboxed execution}
\label{sec:sandbox}
Agents synthesize Python code executed in a persistent Jupyter kernel within an isolated filesystem sandbox. This design ensures three critical properties. First, \textit{State Persistence}: unlike stateless function calls, the kernel maintains active memory of variables, loaded datasets, and imported libraries, allowing iterative multi-step analysis without data reloading. Second, \textit{Session Isolation}: each user session initializes a unique workspace identified by a universally unique identifier (UUID), with all data retrieval and artifact generation confined to this directory to prevent cross-session contamination. Third, \textit{Self-Correction}: the kernel captures standard error outputs (tracebacks) and feeds them back to the generating agent, which parses the traceback, refactors the code, and re-executes.

\subsubsection{Cross-model error escalation (Wise Agent)}
\label{sec:wise_agent}
When a worker agent fails to resolve an error after a configurable number of retry cycles, the system escalates the task to a secondary agent (the \emph{Wise Agent}) backed by a different foundation model. In the current deployment, the primary pipeline runs on GPT-5.2 \citep{openai2025gpt52}, while the Wise Agent runs on Claude Opus~4.6 \citep{anthropic2025claude4}. The escalation transfers the full execution context (code, traceback, and prior repair attempts) to the Wise Agent, which generates an independent fix. This cross-model fallback exploits the observation that different LLMs exhibit complementary failure modes: an error pattern that causes one model to loop often lies outside the failure distribution of another. In practice, this mechanism resolved deadlocked error cycles that single-model retries could not break.

\subsubsection{Visual quality control}
\label{sec:reflection}
Visual outputs are subjected to a reflexive quality-control loop that intercepts every generated figure before it is surfaced to the user. Once a plot is saved to the sandbox, the image is passed to a vision-enabled language model acting as a structured critic. The critic evaluates the figure against a ten-point rubric spanning five dimensions: axis labelling and units, legend completeness, adherence to domain conventions (e.g., inverted depth axes for oceanographic profiles, standard color palettes for temperature fields), absence of overlapping text, and overall readability at publication scale. If the composite score falls below a strict threshold, the critic returns structured, actionable feedback specifying each deficiency. This feedback is appended to the generating agent's context, triggering a targeted code refactoring cycle. The loop repeats until the score exceeds the threshold or a maximum of five iterations is reached.

\section*{Data Availability}
\begin{sloppypar}
All datasets analyzed in this study are publicly available through the
PANGAEA data repository. Specific datasets are referenced by DOI in the
corresponding scenario descriptions: Weddell Sea microplastics
(\href{https://doi.org/10.1594/PANGAEA.941076}{10.1594/PANGAEA.941076}),
MOSAiC PS122/1 navigation
(\href{https://doi.org/10.1594/PANGAEA.935221}{10.1594/PANGAEA.935221}),
North Atlantic jellyfish survey
(\href{https://doi.org/10.1594/PANGAEA.829702}{10.1594/PANGAEA.829702}),
and HAUSGARTEN moorings (PANGAEA DOIs 845608--845617).
Climate reanalysis data were accessed from the Copernicus Marine Service
(GLORYS12V1) and the ERA5 reanalysis accessed via the Arraylake
cloud-native data platform (Earthmover, \url{https://earthmover.io}).
The benchmark query dataset (100 queries, Supplementary Note~5)
is available at \url{https://github.com/CliDyn/pangaeaGPT}.
\end{sloppypar}

\section*{Code Availability}
\begin{sloppypar}
The PANGAEA-GPT source code, agent configurations, and benchmark
evaluation scripts are available at
\url{https://github.com/CliDyn/pangaeaGPT}.
\end{sloppypar}

\section*{Author Contributions}
D.P.\ conceived and designed the system, developed the multi-agent
architecture, conducted all experiments, and wrote the manuscript.
I.K.\ provided domain expertise in physical oceanography, guided the
design of the validation scenarios, and reviewed the scientific
interpretation of the results.
B.S.\ contributed to the system design and assisted with the integration
of the PANGAEA Elasticsearch API and metadata harvesting protocols.
A.A.J.\ contributed to the ecological use-case design, provided expertise
in biodiversity metrics, and assisted with the curation of the benchmark
query dataset.
N.K.\ supervised the project, provided strategic guidance on the system
architecture and its integration with the PANGAEA infrastructure, and
contributed to manuscript revision.
T.J.\ provided institutional support, strategic oversight, and
contributed to manuscript revision.
All authors reviewed and approved the final manuscript.

\section*{Competing Interests}
The authors declare no competing interests.

\section*{Acknowledgements}
This work was supported by the Helmholtz Association and the Federal Ministry of Education and Research (BMBF) for supporting the DataHub Initiative of the Research Field Earth and Environment; the European Union's Destination Earth Initiative and relates to tasks entrusted by the European Union to the European Centre for Medium-Range Weather Forecasts implementing part of this Initiative with funding by the European Union; and Projects S1: Diagnosis and Metrics in Climate Models of the Collaborative Research Centre TRR 181 ``Energy Transfer in Atmosphere and Ocean,'' funded by the Deutsche Forschungsgemeinschaft (DFG, German Research Foundation, project no.\ 274762653). Views and opinions expressed are those of the authors only and do not necessarily reflect those of the European Union or the European Climate Infrastructure and Environment Executive Agency (CINEA). Neither the European Union nor the granting authority can be held responsible for them.

\bibliographystyle{plainnat}
\bibliography{references}

\end{document}


\maketitle

\renewcommand{\thetable}{S\arabic{table}}
\setcounter{table}{0}

\section*{Supplementary Note 1: Scenario~1 extended protocol}

This section provides the complete methodological detail for Scenario~1 (Weddell Sea microplastic exploration), including verbatim user prompts, the agent task decomposition, autonomous error-recovery episodes, and the visualization refinement cycle. The condensed main-text account (Scenario~1 in the main text) should be read alongside Supplementary Table~S1 below.

\subsection*{1.1\quad Data and scientific context}
The user loaded a published dataset of sea-surface microplastic concentrations collected by Manta trawl during two RV~Polarstern expeditions to the Weddell Sea (PS111, austral summer 2018; PS117, austral summer 2019; Dataset DOI: \href{https://doi.org/10.1594/PANGAEA.941076}{10.1594/PANGAEA.941076}). The dataset comprises $N=34$ sampling stations spanning 78$^\circ$S--60$^\circ$S and 57$^\circ$W--12$^\circ$E, with microplastic abundance reported as particle counts per unit area (items\,km$^{-2}$).

\subsection*{1.2\quad User prompts and task decomposition}
The workflow was driven by three user prompts. The first directed the system to retrieve Copernicus Marine surface currents, calculate mean current speed over austral summers 2018--2019, and create a regional map. The Supervisor decomposed this into two sub-tasks: (i)~data retrieval and temporal averaging, routed to the Oceanographer Agent; and (ii)~cartographic rendering, routed to the Visualization Agent. The second prompt requested a minor aesthetic refinement (40\% transparency). The third prompt directed the statistical analysis (Spearman correlation, Mann--Whitney $U$ test, log-linear regression plot).

\subsection*{1.3\quad Self-correction episodes}
Two autonomous error-recovery events occurred during execution:

\paragraph{Episode 1: Copernicus depth coordinate error (steps~4--5).}
The first API request specified \texttt{minimum\_depth=0.0}, which the Copernicus Marine Service rejected because the depth axis begins at 0.494\,m. The agent parsed the error message, extracted the minimum valid depth coordinate, corrected the parameter, and resubmitted both requests successfully.

\paragraph{Episode 2: Visualization layout failure (step~8).}
The reflexive quality control loop rated the initial regional map at 3/10. The Visualization Agent autonomously refactored the figure layout, achieving a revised score of 9/10.

\subsection*{1.4\quad Statistical analysis protocol}
The DataFrame Agent co-located each of the 34 sampling stations with the nearest grid point of the mean current speed field, producing a matched dataset of $N=21$ stations after filtering. Statistics: Spearman $\rho = -0.466$ ($p = 0.033$); Mann--Whitney $U = 88.0$ ($p = 0.022$); OLS regression on $\log_{10}$(abundance): slope $= -4.03$, intercept $= 3.58$.

{\footnotesize
\renewcommand{\arraystretch}{1.3}
\begin{longtable}{p{0.03\linewidth} p{0.13\linewidth} p{0.17\linewidth} p{0.57\linewidth}}
    \caption{Annotated execution log for Scenario~1 (Weddell Sea microplastic exploration). Bold text indicates autonomous agent decisions. The workflow comprises three user prompts (Steps~1, 9, 11) and eleven autonomous agent actions, including one API error-recovery episode (Steps~4--5) and one reflexive visualization refinement cycle (Step~8).}
    \label{tab:supp_log} \\
    \hline
    \textbf{\#} & \textbf{Actor} & \textbf{Action} & \textbf{Content / Reasoning} \\
    \hline
    \endfirsthead
    \hline
    \textbf{\#} & \textbf{Actor} & \textbf{Action} & \textbf{Content / Reasoning} \\
    \hline
    \endhead
    \hline
    \endfoot
    1 & User & Prompt &
    ``Retrieve Copernicus Marine surface currents (\texttt{uo}, \texttt{vo}) for [78$^\circ$S--60$^\circ$S, 57$^\circ$W--12$^\circ$E]. Calculate mean current speed averaged over austral summers 2018--2019. Create a regional map with current speed heatmap, streamlines, and sampling stations sized by microplastic abundance.'' \\
    \hline
    2 & Supervisor & Planning &
    Decomposes into two sub-tasks: (i)~data retrieval and processing $\rightarrow$ Oceanographer Agent; (ii)~cartographic visualization $\rightarrow$ Visualization Agent. \\
    \hline
    3 & Oceanographer & \textbf{Temporal inference} &
    \textbf{Autonomously} parses the \texttt{Date/Time} column of the observational dataset and determines the sampling campaigns span January--February 2018 and January--February 2019. Constructs two separate API requests with matching temporal bounds. \\
    \hline
    4 & Oceanographer & API call (\textbf{fail}) &
    First Copernicus Marine request specifies \texttt{minimum\_depth=0.0}. Service rejects the request: depth axis begins at 0.494\,m. \\
    \hline
    5 & Oceanographer & \textbf{Self-correction} &
    Parses the error message, extracts the minimum valid depth coordinate (0.494\,m), and resubmits. Both requests (2018, 2019) succeed. Downloads daily $u$, $v$ fields for the specified domain. \\
    \hline
    6 & Oceanographer & Code execution &
    Computes temporal mean of daily velocity fields across both summers using \texttt{xarray}. Derives current speed as $\sqrt{u^2 + v^2}$. Saves result as NetCDF (\texttt{mean\_surface\_currents\_JF\_2018\_2019\_Weddell.nc}). \\
    \hline
    7 & Visualization & Code execution &
    Generates map using South Polar Stereographic projection (\texttt{cartopy}). Overlays streamlines on current speed heatmap (\texttt{cmocean.deep} colormap). Plots stations as circles scaled by abundance, color-coded by concentration (\texttt{magma} colormap). \\
    \hline
    8 & Visualization & \textbf{Reflection loop} &
    Reflexive quality control scores initial figure at \textbf{3/10}: ``legend overlaps with map area, significantly impacts readability.'' Agent refactors layout: repositions colorbars into dedicated axes, relocates size legend outside the plot area, adjusts subplot margins. Revised version scores \textbf{9/10}: ``no overlapping elements, effective data communication.'' \\
    \hline
    9 & User & Refinement &
    ``Make the same figure, but draw circles with microplastics that are 40\% transparent.'' \\
    \hline
    10 & Visualization & Code execution &
    Re-executes the plotting script with a single parameter change (\texttt{alpha=0.6}). \\
    \hline
    11 & User & Analysis request &
    ``Calculate the Spearman Rank Correlation between current speed and microplastic abundance. Perform a Mann--Whitney $U$ test splitting stations into Low vs.\ High Kinetic Energy groups at the median speed. Generate a log-linear regression plot with 95\% CI and annotated statistics.'' \\
    \hline
    12 & DataFrame & Co-location &
    Loads the mean current speed field. For each of the 34 sampling stations, extracts the nearest-grid-point current speed via \texttt{xarray} vectorized selection. Filters to $N=21$ stations with positive abundance and valid current speed. \\
    \hline
    13 & DataFrame & Statistics &
    Computes Spearman $\rho=-0.466$ ($p=0.033$). Splits stations at median speed (0.034\,m\,s$^{-1}$): $n_{\mathrm{low}}=11$, $n_{\mathrm{high}}=10$. Mann--Whitney $U=88.0$ ($p=0.022$). Fits OLS regression on $\log_{10}$(abundance): slope $=-4.03$, intercept $=3.58$. \\
    \hline
    14 & Visualization & Code execution &
    Generates scatter plot with logarithmic $y$-axis, regression line, 95\% CI band, and statistical annotations. Reflexive quality control scores 8/10 on first attempt. \\
    \hline
    \end{longtable}
}

\section*{Supplementary Note 2: Scenario~2 extended protocol}

This section documents the complete interaction sequence for Scenario~2 (Copernicus/HAUSGARTEN mooring validation), including data characteristics, co-location methodology, and all autonomous error-recovery episodes.

\subsection*{2.1\quad Data and scientific context}
The user loaded ten published mooring temperature datasets from the LTER HAUSGARTEN observatory in the eastern Fram Strait (PANGAEA DOIs 10.1594/PANGAEA.845608 through 10.1594/PANGAEA.845617). Together the datasets contain 340,022 records spanning 2006--2012, with instrument depths from 81--2,737\,m.

\subsection*{2.2\quad Self-correction episodes}
One autonomous error-recovery event occurred:

\paragraph{Episode 1: Datetime type mismatch (Oceanographer agent).}
\texttt{AttributeError}: \texttt{numpy.datetime64} has no \texttt{.date()} method. The agent refactored to \texttt{pd.Timestamp.normalize()} for consistent datetime keys.

\subsection*{2.3\quad Validation statistics}
Overall ($N=135{,}678$): Bias $+0.35\,^\circ$C, RMSE $1.09\,^\circ$C, Pearson $r=0.31$.

\begin{table}[H]
\centering
\footnotesize
\begin{tabular}{lcccc}
\hline
Depth bin & $N$ & Bias ($^\circ$C) & RMSE ($^\circ$C) & $r$ \\
\hline
50--100\,m  &  7,738 & $+0.20$ & 0.97 & $-0.05$ \\
100--200\,m & 47,119 & $+0.42$ & 1.06 & $+0.11$ \\
200--500\,m & 80,821 & $+0.32$ & 1.12 & $+0.42$ \\
\hline
\end{tabular}
\caption{Depth-binned validation statistics for GLORYS12V1 potential temperature against HAUSGARTEN mooring observations (0--500\,m, 2006--2012).}
\label{tab:depth_bin_validation}
\end{table}

{\footnotesize
\renewcommand{\arraystretch}{1.3}
\begin{longtable}{p{0.03\linewidth} p{0.13\linewidth} p{0.17\linewidth} p{0.57\linewidth}}
    \caption{Annotated execution log for Scenario~2 (HAUSGARTEN mooring validation). Bold text indicates autonomous agent decisions. The workflow comprises three user prompts (Steps~1, 9, 11) and sixteen autonomous agent actions, including one error-recovery episode (Steps~15--16, datetime type mismatch) and two reflexive visualization quality control cycles (Steps~6--8, 19).}
    \label{tab:supp_log_s2} \\
    \hline
    \textbf{\#} & \textbf{Actor} & \textbf{Action} & \textbf{Content / Reasoning} \\
    \hline
    \endfirsthead
    \hline
    \textbf{\#} & \textbf{Actor} & \textbf{Action} & \textbf{Content / Reasoning} \\
    \hline
    \endhead
    \hline
    \endfoot
    1 & User & Prompt &
    ``Generate a plot visualizing the distribution of instrument depths across all loaded mooring datasets.'' \\
    \hline
    2 & Supervisor & Planning &
    Decomposes into two sub-tasks: (i)~multi-dataset ingestion and depth extraction $\rightarrow$ Visualization Agent (with inline data wrangling); (ii)~exploratory depth-distribution figure $\rightarrow$ Visualization Agent. \\
    \hline
    3 & Visualization & \textbf{Data profiling} &
    Loads all 10 CSV files and inspects column names. Identifies \texttt{Depth water} as the depth column across all datasets. Notes that three datasets (FEVI21, FEVI22, FEVI23) contain an additional \texttt{Sal} column (14 columns vs.\ 13). Records total row counts: 340,022 records across 10 moorings. \\
    \hline
    4 & Visualization & \textbf{Data engineering} &
    Computes per-instrument unique depths to avoid time-repeated values: groups by \texttt{Gear~ID} within each event and takes the median of \texttt{Depth~water}. Produces a consolidated table of $N=35$ unique instrument depths across 10 moorings, spanning 81--2,737\,m. \\
    \hline
    5 & Visualization & Code execution &
    Generates a two-panel figure: (top) overall histogram with KDE of the 35 unique instrument depths; (bottom) violin + box + strip plot faceted by mooring event, ordered by median depth. Annotates per-mooring instrument counts ($n=3$--$5$). \\
    \hline
    6 & Visualization & \textbf{Reflection loop} &
    Reflexive quality control scores the initial figure at \textbf{7/10}: ``Axis labels slightly small; histogram x-axis extends into negative values; font sizes could be increased for readability.'' \\
    \hline
    7 & Visualization & \textbf{Self-correction} &
    Refactors the figure: increases font sizes to 14\,pt for axis labels and 16\,pt for titles; clips KDE at zero (\texttt{clip=(0, None)}); sets \texttt{xlim(left=0)}; fixes seaborn \texttt{FutureWarning} by explicitly assigning \texttt{hue} parameter with \texttt{legend=False}. \\
    \hline
    8 & Visualization & \textbf{Reflection loop} &
    Revised figure scores \textbf{9/10}: ``All labels clear, no overlaps, scientific conventions followed, depth represented correctly on the Y-axis.'' Final plot accepted. \\
    \hline
    9 & User & Prompt &
    ``Create a consolidated dataset by merging all loaded mooring records from the upper ocean, specifically between 0 and 500 meters depth. Preserve all original data columns and save to results.'' \\
    \hline
    10 & DataFrame & Code execution &
    Iterates over all 10 datasets: coerces \texttt{Depth~water} to numeric, filters to $0 \leq \texttt{Depth~water} \leq 500$\,m, appends a \texttt{mooring\_id} column, and concatenates. Produces merged table: 136,183 rows $\times$ 15 columns. Saves \texttt{merged\_moorings\_upper\_ocean\_0\_500m.csv}. \\
    \hline
    11 & User & Prompt &
    ``Validate Copernicus global reanalysis data against in-situ mooring observations (first 500\,m). Determine spatio-temporal bounds, retrieve daily \texttt{thetao}, co-locate with moorings, calculate Bias/RMSE/Correlation, and generate a scatter plot. Do it in one step with the Oceanographer Agent.'' \\
    \hline
    12 & Oceanographer & Code execution &
    Loads all 10 mooring datasets, merges (340,022 rows), parses \texttt{Date/Time} to UTC datetime, coerces numeric columns. Filters to 0--500\,m with valid temperature, time, latitude, and longitude: $N=135{,}678$ records. Extracts bounds: time 2006-09-05 to 2012-07-23; lat 78.61$^\circ$N--79.74$^\circ$N; lon 4.33$^\circ$E--5.16$^\circ$E; depth 81--500\,m. \\
    \hline
    13 & Oceanographer & API call &
    Submits Copernicus Marine Service request for \texttt{thetao} from GLORYS12V1 (\texttt{cmems\_mod\_glo\_phy\_my\_0.083deg\_P1D-m}) with the extracted spatio-temporal bounds. Receives NetCDF: 2,149 daily time steps $\times$ 31 depth levels $\times$ 13 latitude $\times$ 10 longitude grid points. \\
    \hline
    14 & Oceanographer & Code execution &
    Implements 4D nearest-grid-point co-location: builds a date$\rightarrow$index lookup dictionary from the model time axis; computes nearest-neighbor indices for latitude, longitude, and depth using vectorized operations. Extracts model \texttt{thetao} values at all 135,678 observation points via NumPy advanced indexing. \\
    \hline
    15 & Oceanographer & Code execution (\textbf{fail}) &
    \texttt{AttributeError}: \texttt{numpy.datetime64} object has no attribute \texttt{.date()} during time-index mapping. The agent's dictionary comprehension attempted to call \texttt{.date()} on \texttt{numpy.datetime64} elements returned by \texttt{pd.to\_datetime().values}. \\
    \hline
    16 & Oceanographer & \textbf{Self-correction} &
    Identifies the type mismatch between NumPy datetime64 and Python native datetime. Refactors to use \texttt{pd.Timestamp.normalize()} for both the model time axis and the observation dates, producing consistent \texttt{Timestamp} keys. Re-executes co-location: 100\% valid fraction (all 135,678 matchups successful). \\
    \hline
    17 & Oceanographer & Statistics &
    Computes overall validation statistics: Bias $= +0.349\,^\circ$C, RMSE $= 1.093\,^\circ$C, Pearson $r = 0.310$. Computes depth-bin statistics: 50--100\,m ($n=7{,}738$, Bias $+0.20$, RMSE 0.97, $r=-0.05$); 100--200\,m ($n=47{,}119$, Bias $+0.42$, RMSE 1.06, $r=+0.11$); 200--500\,m ($n=80{,}821$, Bias $+0.32$, RMSE 1.12, $r=+0.42$). \\
    \hline
    18 & Oceanographer & Code execution &
    Generates validation scatter plot: mooring temperature vs.\ GLORYS12V1 \texttt{thetao}, color-coded by observation depth (\texttt{viridis} colormap), with 1:1 reference line and annotation box displaying $N$, Bias, RMSE, $r$. Subsamples to 50,000 points for rendering. \\
    \hline
    19 & Oceanographer & \textbf{Reflection loop} &
    Reflexive quality control scores the scatter plot at \textbf{8/10}: ``Professionally crafted, effective communication, minor improvements possible in legend font size.'' No further correction required. \\
    \hline
    \end{longtable}
}

\section*{Supplementary Note 3: Scenario~3 extended protocol}

This section documents the complete interaction sequence for Scenario~3 (ERA5/MOSAiC Lagrangian validation), including verbatim user prompts, data characteristics, the resource-efficient retrieval strategy, and all autonomous error-recovery episodes.

\subsection*{3.1\quad Data and scientific context}
The user loaded a published minute-resolution underway dataset from the first leg of the MOSAiC expedition (PS122/1; Dataset DOI: \href{https://doi.org/10.1594/PANGAEA.935221}{10.1594/PANGAEA.935221}). The dataset contains 120,719 records across 22 columns, with temporal coverage spanning 2019-09-20 to 2019-12-13 and positions ranging from 69.68$^\circ$N to 86.60$^\circ$N and 19.00$^\circ$E to 138.14$^\circ$E.

\subsection*{3.2\quad User prompts}
Two user prompts drove the workflow: (1)~validate ERA5 against ship measurements while minimizing download, and (2)~generate a Wind Rose diagram with 16 cardinal bins and Beaufort color coding.

\subsection*{3.3\quad Self-correction episodes}
Two autonomous error-recovery events occurred:

\paragraph{Episode 1: ERA5 bounding-box edge effect.}
After initial interpolation, 83 of 2,013 co-located points returned NaN. The agent expanded the retrieval domain by ${\sim}$0.3$^\circ$ and re-downloaded the subset, producing zero NaN values.

\paragraph{Episode 2: Wind rose bin-label mismatch.}
The Visualization Agent defined 13 Beaufort speed-bin edges but provided 13 labels (12 required). The agent parsed the \texttt{ValueError} traceback, removed the extraneous label, and re-executed successfully.

{\footnotesize
\renewcommand{\arraystretch}{1.3}
\begin{longtable}{p{0.03\linewidth} p{0.13\linewidth} p{0.17\linewidth} p{0.57\linewidth}}
    \caption{Annotated execution log for Scenario~3 (ERA5/MOSAiC Lagrangian validation). Bold text indicates autonomous agent decisions. The workflow comprises two user prompts (Steps~1, 11) and fourteen autonomous agent actions, including two error-recovery episodes (Steps~6--8, bounding-box edge effect; Steps~13--14, bin-label mismatch) and two reflexive visualization quality control cycles (Steps~10, 16).}
    \label{tab:supp_log_s3} \\
    \hline
    \textbf{\#} & \textbf{Actor} & \textbf{Action} & \textbf{Content / Reasoning} \\
    \hline
    \endfirsthead
    \hline
    \textbf{\#} & \textbf{Actor} & \textbf{Action} & \textbf{Content / Reasoning} \\
    \hline
    \endhead
    \hline
    \endfoot
    1 & User & Prompt &
    ``Validate ERA5 reanalysis against the ship's measurements while minimizing data download size. Calculate the exact spatial bounding box and time range from the ship's data. Retrieve ERA5 \texttt{2m\_temperature} only for that specific region. Interpolate the grid to the ship's hourly positions, convert to Celsius. Create a time-series plot comparing in-situ air temperature (column \texttt{TTT}) in blue against ERA5 temperature in red.'' \\
    \hline
    2 & Supervisor & Planning &
    Decomposes into three sub-tasks: (i)~spatiotemporal bounds extraction and hourly resampling $\rightarrow$ DataFrame Agent; (ii)~ERA5 cloud-native retrieval and Lagrangian interpolation $\rightarrow$ Oceanographer Agent; (iii)~time-series visualization $\rightarrow$ Visualization Agent. \\
    \hline
    3 & DataFrame & \textbf{Temporal inference} &
    Parses the \texttt{Date/Time} column and computes spatiotemporal bounds: bounding box [18.99$^\circ$E--138.14$^\circ$E, 69.68$^\circ$N--86.60$^\circ$N], time range 2019-09-20 to 2019-12-13. Resamples to hourly resolution (last observation per hour), producing $N=2{,}013$ hourly positions. Saves \texttt{hourly\_ship\_table.csv}. \\
    \hline
    4 & Oceanographer & API call &
    Submits ERA5 cloud-native retrieval request for \texttt{2m\_temperature} with bounding box [18.99$^\circ$E--138.14$^\circ$E, 69.68$^\circ$N--86.60$^\circ$N] and temporal range matching the cruise. Receives Zarr store: 2,013 time steps $\times$ 68 latitude $\times$ 477 longitude (${\sim}$261\,MB). \\
    \hline
    5 & Oceanographer & Code execution &
    Interpolates ERA5 \texttt{2m\_temperature} field to $N=2{,}013$ hourly ship positions using trilinear interpolation (\texttt{xarray.interp}, method=\texttt{linear}). Converts from Kelvin to Celsius. \\
    \hline
    6 & Oceanographer & \textbf{Quality check (fail)} &
    Detects 83 of 2,013 interpolated values returned \texttt{NaN}. Diagnoses the cause: ship positions near the bounding-box edge fall outside the ERA5 grid domain (ERA5 grid starts at 69.75$^\circ$N / 19.00$^\circ$E, but ship positions extend to 69.68$^\circ$N / 18.99$^\circ$E). \\
    \hline
    7 & Oceanographer & \textbf{Self-correction} &
    Expands the retrieval domain by ${\sim}$0.3$^\circ$ in all directions (69.4$^\circ$N--86.9$^\circ$N, 18.7$^\circ$E--138.45$^\circ$E). Re-submits the ERA5 API request. Receives expanded Zarr store: 70 latitude $\times$ 479 longitude grid. \\
    \hline
    8 & Oceanographer & Code execution &
    Re-interpolates using the expanded grid. Verifies: 0 NaN values across all 2,013 co-located points. Saves merged table \texttt{ship\_hourly\_TTT\_vs\_ERA5\_t2m.csv}. \\
    \hline
    9 & Visualization & Code execution &
    Generates dual-line time-series plot: ship in-situ TTT (blue) vs.\ ERA5 interpolated 2\,m temperature (red). Uses \texttt{seaborn-whitegrid} style, 300\,DPI. \\
    \hline
    10 & Visualization & \textbf{Reflection loop} &
    Reflexive quality control scores the time-series figure at \textbf{9/10}: ``Axis labels clearly visible, good contrast, no overlapping elements.'' No further correction required. \\
    \hline
    11 & User & Prompt &
    ``Generate a Wind Rose diagram to visualize the prevailing wind patterns for the whole cruise track. Split the wind directions into 16 cardinal bins and color-code the wedges based on the Beaufort Wind Force scale.'' \\
    \hline
    12 & Visualization & Code execution &
    Loads raw dataset. Identifies true wind columns: \texttt{dd} (direction, $^\circ$) and \texttt{ff} (speed, m\,s$^{-1}$). Filters to $n=120{,}616$ valid observations. \\
    \hline
    13 & Visualization & Code execution (\textbf{fail}) &
    Defines 13 Beaufort speed-bin edges (\texttt{[0, 0.5, 1.6, \ldots, 28.5, $\infty$]}) but provides 13 category labels. \texttt{pd.cut} raises \texttt{ValueError}: ``Bin labels must be one fewer than the number of bin edges.'' \\
    \hline
    14 & Visualization & \textbf{Self-correction} &
    Parses the traceback. Recognizes the off-by-one mismatch: 13 edges require 12 labels. Removes the extraneous ``Hurricane (12)'' label (not represented in data; max speed = 23\,m\,s$^{-1}$). Re-executes \texttt{pd.cut} successfully. Constructs 16-sector $\times$ 12-category frequency table. \\
    \hline
    15 & Visualization & Code execution &
    Generates stacked polar bar chart (wind rose) using South Polar Stereographic convention (0$^\circ$ = North, clockwise). Applies \texttt{viridis} colormap across Beaufort categories. Exports PNG (400\,DPI) and SVG. \\
    \hline
    16 & Visualization & \textbf{Reflection loop} &
    Reflexive quality control scores the wind rose at \textbf{8/10}: ``Clear, legible, standard meteorological convention adhered to.'' No further correction required. \\
    \hline
    \end{longtable}
}

\section*{Supplementary Note 4: Scenario~4 extended protocol}

This section documents the complete interaction sequence for Scenario~4 (Biodiversity and Water Mass Coupling), including verbatim user prompts, statistical outputs, and autonomous error-recovery episodes.

\subsection*{4.1\quad Data and scientific context}
The user loaded a biological dataset of jellyfish abundance collected during the RV~\textit{G.O.\ Sars} cruise (May 2013) in the North Atlantic (DOI: \href{https://doi.org/10.1594/PANGAEA.829702}{10.1594/PANGAEA.829702}). The dataset comprises 94 sampling stations.

\subsection*{4.2\quad User prompts}
Three progressive prompts drove the analysis: (1)~divide stations into Eastern/Western transects at $-20^\circ$ longitude; (2)~calculate Shannon-Wiener Index and generate a violin plot; (3)~retrieve Copernicus Marine $\theta$/$S$ data and create a T-S diagram colored by \textit{Aglantha digitale} abundance.

\subsection*{4.3\quad Self-correction episodes}
Two autonomous error-recovery events occurred:

\paragraph{Episode 1: Mathematical safety.}
During Shannon Index calculation, the agent encountered a \texttt{RuntimeWarning: divide by zero}. It refactored the code to apply a mask (\texttt{where(p>0)}).

\paragraph{Episode 2: Reflexive visual refinement.}
The initial violin plot scored 3/10. The agent iteratively refactored across five cycles (3~$\rightarrow$~4~$\rightarrow$~4~$\rightarrow$~5~$\rightarrow$~9) until scientific standards were met.

\subsection*{4.4\quad Statistical summary}
Welch's t-test: $t=-3.39$, $p=0.0014$. Mean $H'_{\mathrm{West}}=0.81$ ($n=34$, Irminger Sea), $H'_{\mathrm{East}}=0.35$ ($n=60$, Norwegian Sea).

{\footnotesize
\renewcommand{\arraystretch}{1.3}
\begin{longtable}{p{0.03\linewidth} p{0.13\linewidth} p{0.17\linewidth} p{0.57\linewidth}}
    \caption{Annotated execution log for Scenario~4 (G.O.~Sars jellyfish transect analysis with Copernicus validation). Bold text indicates autonomous agent decisions. The workflow comprises three user prompts (Steps~1, 6, 20) and twenty-three autonomous agent actions, including two error-recovery episodes (Steps~7--8, divide-by-zero in log; Steps~11--19, extended reflexive visualization quality control with five iterations, score progression: 3$\rightarrow$4$\rightarrow$4$\rightarrow$5$\rightarrow$9/10), and two single-pass reflection cycles (Steps~5, 26).}
    \label{tab:supp_log_s4} \\
    \hline
    \textbf{\#} & \textbf{Actor} & \textbf{Action} & \textbf{Content / Reasoning} \\
    \hline
    \endfirsthead
    \hline
    \textbf{\#} & \textbf{Actor} & \textbf{Action} & \textbf{Content / Reasoning} \\
    \hline
    \endhead
    \hline
    \endfoot
    \multicolumn{4}{l}{\textit{Phase~1: Geographical transect division}} \\
    \hline
    1 & User & Prompt &
    ``Divide the sampling stations into two geographical transects based on longitude to separate the Norwegian Sea leg from the Irminger Sea leg. Create two new dataframes: `Eastern Transect' and `Western Transect' for stations, use $-20^{\circ}$ longitude as a border. Visualize these transects at the end.'' \\
    \hline
    2 & DataFrame & \textbf{Data engineering} &
    Identifies \texttt{Longitude} and \texttt{Event} columns. Splits at $-20^{\circ}$\,E: \textbf{Eastern Transect} (60~rows, 18~unique stations) and \textbf{Western Transect} (34~rows, 11~unique stations). \\
    \hline
    3 & DataFrame & Code execution &
    Generates lon/lat scatter plot with cruise track, dashed border at $-20^{\circ}$, and ``Irminger Sea leg'' / ``Norwegian Sea leg'' annotations. Saves \texttt{transects\_map\_lonlat\_border\_-20.png}. \\
    \hline
    4 & Visualization & Code execution &
    Parses \texttt{Date/Time}, splits transects, sorts by time. Regenerates the transect map with connected cruise-track lines, styled scatter points, and text annotations. \\
    \hline
    5 & Visualization & \textbf{Reflection loop} &
    Reflexive quality control scores the transect map at \textbf{8/10}: ``Axis labels clear, colors distinguishable, no overlapping elements.'' Accepted without further correction. \\
    \hline
    \multicolumn{4}{l}{\textit{Phase~2: Shannon--Wiener diversity analysis}} \\
    \hline
    6 & User & Prompt &
    ``Calculate the Shannon--Wiener Diversity Index for every station using all available taxa columns. Then, create a split violin plot (or box plot with jitter points) to statistically compare the biodiversity levels between the `Eastern' and `Western' transects.'' \\
    \hline
    7 & DataFrame & \textbf{Ecological computation} &
    Computes station-level Shannon--Wiener $H' = -\sum p_i \ln p_i$ across 64~taxa. Encounters \texttt{RuntimeWarning: divide by zero in log} from zero-abundance proportions. \\
    \hline
    8 & DataFrame & \textbf{Self-correction} &
    Refactors computation: converts proportions to NumPy array, applies \texttt{np.where(p>0, p*np.log(p), 0.0)} to avoid log(0). Verifies: minimum $H'=0$ (no negative values). \\
    \hline
    9 & DataFrame & \textbf{Statistical inference} &
    Computes inferential statistics: Welch's $t = -3.39$, $p = 0.0014$; Mann--Whitney $U = 638$, $p = 0.0019$; Cohen's $d = -0.81$; Hedges' $g = -0.81$; rank-biserial $r = 0.37$. Western transect: $\bar{H}' = 0.81 \pm 0.71$ vs.\ Eastern: $\bar{H}' = 0.35 \pm 0.47$. \\
    \hline
    10 & Visualization & Code execution &
    Generates violin + jitter-strip plot (seaborn) with quartile inner lines, $n$-annotations, and significance bracket with Welch $p$-value. \\
    \hline
    11 & Visualization & \textbf{Reflection loop (iter.~1)} &
    Quality control scores \textbf{3/10}: ``Critical failure: $p$-value annotation and $n$-labels overlap with violin body and data points.'' \\
    \hline
    12 & Visualization & \textbf{Self-correction} &
    Increases Y-axis headroom (\texttt{ylim} $\times 1.45$), repositions $n$-labels and bracket above violins, adds white text-box backgrounds. \\
    \hline
    13 & Visualization & \textbf{Reflection loop (iter.~2)} &
    Quality control scores \textbf{4/10}: ``Annotation $n$=34 still overlaps with data distribution; Welch annotation overlaps with data plot area.'' \\
    \hline
    14 & Visualization & \textbf{Self-correction} &
    Switches to axes-coordinate annotations (\texttt{transform=ax.transAxes}) to guarantee non-overlap regardless of data range. Places $n$-labels at (0.25, 0.98) and (0.75, 0.98), $p$-value at (0.5, 1.08). Sets \texttt{title pad=28}. \\
    \hline
    15 & Visualization & \textbf{Reflection loop (iter.~3)} &
    Quality control scores \textbf{4/10}: ``Title text overlaps with itself, impacting readability.'' \\
    \hline
    16 & Visualization & \textbf{Self-correction} &
    Increases \texttt{title pad} from 28 to 40 to separate the two-line title from the $p$-value annotation. \\
    \hline
    17 & Visualization & \textbf{Reflection loop (iter.~4)} &
    Quality control scores \textbf{5/10}: ``Y-axis is missing units; critical for scientific accuracy.'' \\
    \hline
    18 & Visualization & \textbf{Self-correction} &
    Appends unit annotation to Y-axis label: ``Shannon diversity, $H'$ (unitless; natural log).'' \\
    \hline
    19 & Visualization & \textbf{Reflection loop (iter.~5)} &
    Quality control scores \textbf{9/10}: ``No critical failures. Text and labels clear, appropriately sized. Conventions met.'' Accepted. \\
    \hline
    \multicolumn{4}{l}{\textit{Phase~3: Copernicus T--S diagram with species overlay}} \\
    \hline
    20 & User & Prompt &
    ``Retrieve Copernicus Marine data for Potential Temperature (`thetao') and Salinity (`so') matching the cruise coordinates and depths. Create a Temperature--Salinity (T--S) Diagram. Color-code the scatter points by the abundance of \textit{Aglantha digitale} (using a log scale).'' \\
    \hline
    21 & Oceanographer & \textbf{Data profiling} &
    Loads cruise data. Parses \texttt{Date/Time} (UTC). Computes depth midpoints from \texttt{Depth top}/\texttt{Depth bot} (range: 12.5--900\,m). Extracts spatiotemporal bounds: 59.38--68.79$^{\circ}$N, $-46.19$--5.07$^{\circ}$E, 2013-05-03 to 2013-05-20. \\
    \hline
    22 & Oceanographer & API call &
    Submits Copernicus Marine retrieval for \texttt{cmems\_mod\_glo\_phy\_my\_0.083deg\_P1D-m} (GLORYS12V1): variables \texttt{thetao}, \texttt{so}; domain [$-50^{\circ}$--$10^{\circ}$E, 58--70$^{\circ}$N, 0--1000\,m]; temporal range 2013-05-01 to 2013-05-22. Receives NetCDF: 22~days $\times$ 35~depths $\times$ 145~lats $\times$ 721~lons (${\sim}$1.3\,GB). \\
    \hline
    23 & Oceanographer & \textbf{Co-location} &
    Implements 4D nearest-grid-point matching using \texttt{xarray.sel(method='nearest')} across time (daily floor), latitude, longitude, and depth. Achieves 94/94 (100\%) valid matchups. Saves \texttt{cruise\_samples\_matched\_cmems\_thetao\_so.csv}. \\
    \hline
    24 & DataFrame & Code execution &
    Loads data, extracts \textit{A.~digitale} abundance column. Computes $\log_{10}(x+1)$ transformation. Produces tidy table: 29~unique sampling events. Saves \texttt{aglantha\_digitale\_abundance\_tidy.csv}. \\
    \hline
    25 & Visualization & Code execution &
    Loads both intermediate CSVs. Inner-joins matched CMEMS data ($n=94$) with \textit{A.~digitale} abundance ($n=29$) on \texttt{Event} key. Generates T--S scatter plot: salinity on $x$-axis, potential temperature on $y$-axis, viridis colormap scaled to $\log_{10}(\text{abundance}+1)$. Adds colorbar, sample-count annotation ($n=94$), and grid. \\
    \hline
    26 & Visualization & \textbf{Reflection loop} &
    Quality control scores \textbf{9/10}: ``High professional quality. Axis labels clear, color gradient effective, conventions correct, no overlapping elements.'' Accepted without further correction. \\
    \hline
    \end{longtable}
}

\section*{Supplementary Note 5: Search benchmarking methodology}
\label{sec:supp_benchmark}

\subsection*{5.1\quad Retrieval architectures under evaluation}
We evaluated three retrieval architectures: (1)~Baseline (standard Elasticsearch BM25 keyword matching); (2)~Simple LLM (GPT-5.2 zero-shot query translation); (3)~Agentic Search (our ReAct-based iterative agent).

\subsection*{5.2\quad Query dataset construction}
We constructed a dataset of 100 natural language queries stratified across five complexity categories (20 queries each):
\begin{itemize}
    \item \textbf{Category~1: Specific Entity Search.} Requests for a known, unique dataset.
    \item \textbf{Category~2: Broad Thematic Search.} General exploration without strict spatiotemporal filters.
    \item \textbf{Category~3: Spatiotemporal Slicing.} Queries requiring translation of semantic locations to coordinates.
    \item \textbf{Category~4: Parameter-Specific.} Requests targeting specific chemical or physical variables.
    \item \textbf{Category~5: Cross-Domain Constraints.} Complex queries combining biological, physical, and geological filters.
\end{itemize}

\subsection*{5.3\quad Evaluation protocol and metrics}
For each query, the top-5 retrieved datasets from each architecture were evaluated programmatically using Gemini 3 Pro Preview (Google DeepMind) as an automated judge, scoring on a scale of 0--10 across five semantic metrics:
\begin{enumerate}
    \item \textbf{Precision / Relevance (M1):} Overall relevance to the scientific question.
    \item \textbf{Spatial / Temporal Accuracy (M2):} Adherence to the requested geographic and temporal constraints.
    \item \textbf{Parameter Coverage (M3):} Whether the datasets contain the specific variables requested.
    \item \textbf{Access / Usability (M4):} Suitability for immediate computational analysis.
    \item \textbf{Noise Reduction (M5):} Ability to filter out false positives.
\end{enumerate}
Final scores per architecture were computed as the arithmetic mean across all 100 queries for each metric.

\begin{table}[ht]
    \centering
    \footnotesize
    \renewcommand{\arraystretch}{1.3}
    \begin{tabular}{p{0.15\linewidth} c p{0.38\linewidth} c c}
    \hline
    \textbf{Scenario} & \textbf{Errors} & \textbf{Error Types} &
    \textbf{VQC Cycles} & \textbf{Recovery} \\
    \hline
    1: Weddell Sea    & 1 & API depth parameter validation & 1 & 1/1 \\
    2: HAUSGARTEN     & 1 & Datetime type mismatch & 2 & 1/1 \\
    3: MOSAiC/ERA5    & 2 & Bounding-box edge effect, bin-label mismatch & 2 & 2/2 \\
    4: Jellyfish T-S  & 1 & Divide-by-zero in $\log$ & 6 & 1/1 \\
    \hline
    \textbf{Total}    & \textbf{5} & & \textbf{11} & \textbf{5/5 (100\%)} \\
    \hline
    \end{tabular}
    \caption{Summary of autonomous error recovery across all validation scenarios. ``Errors'' counts distinct runtime failures resolved via kernel traceback introspection. ``VQC Cycles'' counts the total number of reflexive visual quality-control (VQC) cycles performed, including both correction iterations (score below acceptance threshold) and single-pass evaluations. All errors were resolved without human intervention.}
    \label{tab:error_summary}
\end{table}

\section*{Supplementary Note 6: Qualitative retrieval analysis}
\label{sec:supp_qualitative}

To complement the aggregate benchmark statistics reported in the main text, we present a qualitative comparison of representative queries that expose distinct failure modes in the baseline architectures. Table~S7 compares the top-ranked result returned by each method for four queries selected to span different categories of semantic complexity: temporal reasoning, measurement geometry disambiguation, geographic constraint enforcement, and scientific product-level intent recognition.

\subsection*{6.1\quad Temporal reasoning (Q065)}
\textbf{Query:} ``Salinity and temperature profiles from the Weddell Sea during winter 2013.''

The term ``winter'' in a Southern Hemisphere context maps to June--September, a constraint that requires hemispheric awareness absent from keyword indices. The Baseline returned IceBird Winter altimetry campaigns from 2019 and 2023, matching the keyword ``Winter'' while violating both the year and parameter constraints. The Simple LLM retrieved oceanographic profiles but selected summer cruises (ANT-XXIX/2, PS82), failing to invert the seasonal calendar. The Agentic Search correctly resolved the hemispheric constraint and ranked the ANT-XXIX/6 (AWECS) cruise, conducted June--August 2013, as the top result (Agentic mean: 9.8/10; Baseline mean: 1.8/10).

\subsection*{6.2\quad Source disambiguation (Q059)}
\textbf{Query:} ``Satellite altimetry data for studying sea level variability.''

This query exposes what we term the \emph{validation trap}: archival metadata frequently contains phrases such as ``used to validate satellite altimetry'' in records describing in-situ mooring deployments. Both the Baseline and Simple LLM returned these ground-truth observations, confusing validation context with the product itself. The Agentic Search correctly distinguished between the two categories, prioritizing Level-3/4 gridded sea-surface height anomaly products from CryoSat-2, Envisat, ERS-2, and TOPEX/Poseidon. It was the only method to retrieve the foundational WOCE TOPEX/Poseidon datasets (Agentic mean: 9.6/10; Baseline mean: 1.0/10).

\subsection*{6.3\quad Geographic constraint enforcement (Q080)}
\textbf{Query:} ``Ground-penetrating radar (GPR) data from a polar glacier.''

The Baseline and Simple LLM returned GPR datasets from Alpine glaciers in Austria and Italy, matching ``Glacier'' and ``Radar'' while ignoring the ``polar'' constraint. The Agentic Search strictly enforced the geographic restriction: every result in its top-30 list originated from Antarctica or Greenland. It also demonstrated domain-aware semantic expansion, recognising that ``Radio-Echo Sounding'' (RES) is the standard airborne equivalent of ground-based GPR in polar glaciology, thereby broadening recall without sacrificing precision (Agentic mean: 9.8/10; Baseline mean: 2.8/10).

\subsection*{6.4\quad Product vs.\ raw data intent (Q096)}
\textbf{Query:} ``Coastal erosion rates from historical aerial photographs and LiDAR.''

The Simple LLM and Baseline returned raw aerial imagery subsets (Rettelbach et al.\ 2024, occupying 17--18 positions in each result list), treating the input methodology as the retrieval target rather than the requested scientific product (erosion rates). The Agentic Search prioritised derived products: all top-6 results were datasets quantifying shoreline retreat rates (Yukon, Laptev Sea, Baltic Sea, Portugal), correctly interpreting the user's intent as the output variable rather than the input data source (Agentic mean: 9.0/10; Baseline mean: 2.6/10).

\begin{table}[H]
    \centering
    \footnotesize
    \renewcommand{\arraystretch}{1.4}
    \begin{tabularx}{\linewidth}{|p{0.13\linewidth}|X|X|X|}
    \hline
    \textbf{Query intent} & \textbf{Baseline (Elasticsearch)} & \textbf{Simple LLM} & \textbf{Agentic Search (ours)} \\
    \hline
    \textbf{Q065:} Weddell Sea, Winter 2013 (temporal) &
    \textbf{Failure:} Returns \textit{IceBird Winter 2019/2023} altimetry. Wrong year, wrong parameter. &
    \textbf{Partial:} Returns CTD profiles but from Summer cruises (ANT-XXIX/2, PS82). &
    \textbf{Success:} Returns \textit{ANT-XXIX/6 (AWECS)}, the specific winter cruise (Jun--Aug 2013). \\
    \hline
    \textbf{Q059:} Satellite altimetry (source) &
    \textbf{Failure:} Returns lake water quality and biological transects. &
    \textbf{Validation trap:} Returns in-situ mooring data ``used to validate'' satellites. &
    \textbf{Success:} Returns Level-3/4 gridded SSHA from CryoSat-2, Envisat, TOPEX/Poseidon. \\
    \hline
    \textbf{Q080:} Polar glacier GPR (spatial) &
    \textbf{Geo-drift:} Returns GPR from Alpine glaciers (Austria/Italy). Matches ``Glacier+Radar'' but ignores ``Polar''. &
    \textbf{Mixed:} Returns Antarctic data but includes Alpine mass balance stakes. &
    \textbf{Success:} Strictly polar. All 30 results from Antarctica or Greenland. Expands to RES (Radio-Echo Sounding). \\
    \hline
    \textbf{Q096:} Coastal erosion rates (intent) &
    \textbf{Failure:} Returns vegetation maps and oceanographic bottle data. &
    \textbf{Intent miss:} Returns raw aerial imagery subsets (18/30 results from one campaign). &
    \textbf{Success:} Returns derived shoreline retreat rates (Yukon, Laptev, Baltic, Portugal). \\
    \hline
    \end{tabularx}
    \caption{Head-to-head comparison of the top-ranked search result for four representative complex queries, illustrating qualitative failure modes in baseline methods that are resolved by the agentic architecture. Each query tests a distinct category of semantic reasoning: temporal constraint resolution, source vs.\ validation data disambiguation, geographic filtering, and product-level intent recognition.}
    \label{tab:qualitative_search}
\end{table}

\section*{Supplementary Note 7: Dynamic contextualization pipeline}

The Oceanographer Agent contextualizes local observations within global environmental fields through a three-stage pipeline. First, it extracts the spatiotemporal bounding box of the loaded observational dataset (latitude/longitude extremes, temporal range, depth coverage). Second, it maps the user's scientific request to the appropriate external product: ERA5 for atmospheric variables, GLORYS12V1 (Copernicus Marine Service) for ocean state variables, or satellite-derived products for surface fields such as chlorophyll-$a$ or sea-surface temperature. Third, it constructs a minimal retrieval request, specifying only the required variable, pressure level or depth layer, and spatiotemporal subset, and executes the download via the provider's API. For high-volume atmospheric data, the system accesses the ERA5 reanalysis through the Earthmover Arraylake repository, a cloud-native chunked Zarr store that supports lazy loading via \texttt{xarray} and \texttt{fsspec}. This architecture reduces data transfer from terabytes (full global fields) to the megabyte-scale subsets actually required for the analysis. Once retrieved, the agent performs geometric alignment: for Eulerian comparisons, it co-locates observations to the nearest grid point using the Haversine distance; for Lagrangian platforms (e.g., ship tracks), it performs 4D interpolation along the trajectory in time, latitude, longitude, and depth.